\pdfoutput=1
\documentclass[11pt]{article}

\usepackage[]{acl}

\usepackage{times}
\usepackage{latexsym}
\usepackage{booktabs}
\usepackage{array}

\usepackage[T1]{fontenc}
\usepackage[utf8]{inputenc}

\usepackage{microtype}

\usepackage{inconsolata}

\usepackage{graphicx}
\usepackage{multirow}
\usepackage{pdfpages}

\title{Unveiling the mystery of visual attributes of concrete and abstract concepts: Variability, nearest neighbors, and challenging categories}

\author{
    Tarun Tater$^1$,
    Sabine Schulte im Walde$^1$,
    Diego Frassinelli$^2$
\\ 
$^1$Institute for Natural Language Processing, University of Stuttgart, Germany\\ 
$^2$MaiNLP, Center for Information and Language Processing, LMU Munich, Germany \\ 
\texttt{\{tarun.tater, schulte\}@ims.uni-stuttgart.de} \\
\texttt{frassinelli@cis.lmu.de}
}

\begin{document}

\maketitle

\begin{abstract}

The visual representation of a concept varies significantly depending on its meaning and the context where it occurs; this poses multiple challenges both for vision and multimodal models. 
Our study focuses on concreteness, a well-researched lexical-semantic variable, using it as a case study to examine the variability in visual representations.
We rely on images associated with approximately 1,000 abstract and concrete concepts extracted from two different datasets: Bing and YFCC. Our goals are: (i) evaluate whether visual diversity in the depiction of concepts can reliably distinguish between concrete and abstract concepts; (ii) analyze the variability of visual features across multiple images of the same concept through a nearest neighbor analysis; and (iii) identify challenging factors contributing to this variability by categorizing and annotating images.
Our findings indicate that for classifying images of abstract versus concrete concepts, a combination of basic visual features such as color and texture is more effective than features extracted by more complex models like Vision Transformer (ViT). However, ViTs show better performances in the nearest neighbor analysis, emphasizing the need for a careful selection of visual features when analyzing conceptual variables through modalities other than text.

\end{abstract}

%%%%%%%%%%%%%%%%%%%%%%%%%%%%%%%%%%%%%%%%%%%%%
\section{Introduction}
\label{introduction}

Language and vision play a crucial role for the understanding of the world surrounding us. Among the five senses, vision is considered the primary source of perceptual information for our mental representations when experiencing the real world \citep{BrysbaertEtAl:14, LynottEtAl:20}. Based on these premises, computational studies have leveraged the strong interaction between visual and textual information to uncover the latent relationships between these two modalities and to build richer and more precise representations. 
In most cases, the contribution of these two very different modalities is asymmetric, with the textual modality having a stronger influence on model performance; for example, when investigating the concreteness of a concept, its compositionality, or its semantic representation \citep{BhaskarEtAl:17, Koeper/SchulteImWalde:17b, hewitt-etal-2018-learning}. The exact reasons behind such asymmetry are still unclear, and especially the role of the visual elements has been explored significantly less. 
Therefore, this paper focuses explicitly on the nature and contribution of the visual component. 
To this end, we analyze the different characteristics of concrete and abstract concepts to determine whether and how visual information can help distinguish between them.
Our analysis is particularly important when addressing the complex task of modeling abstract concepts, which often lack a distinctive visual component, unlike their concrete counterpart.
For example, concrete concepts like \texttt{banana} and \texttt{chariot} evoke vivid mental images anchored to objects that are easy to visualize. In contrast, abstract concepts like \texttt{accountability} and \texttt{allegiance} are more challenging and subjective to visualize \citep{paivio1968concreteness, kastner2020estimating}.

\begin{figure*}[!thb]
	\center
    \includegraphics[width=0.98\textwidth]{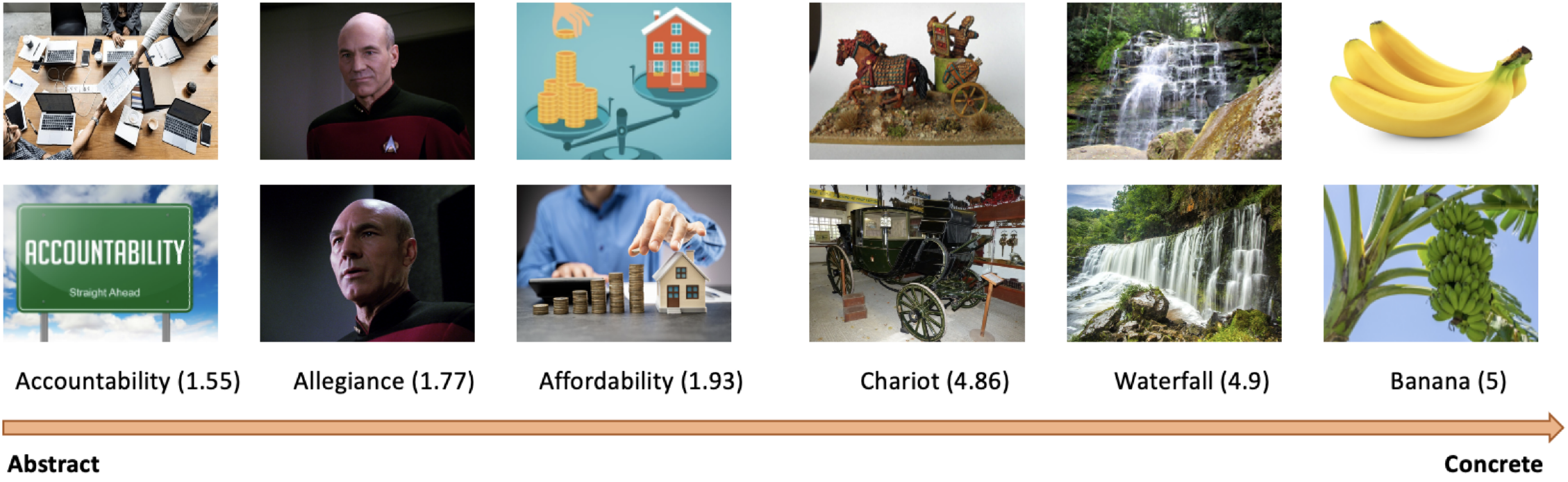}
	\caption{Images of concrete and abstract concepts with varying concreteness ratings on a scale from $1$ (clearly abstract) to $5$ (clearly concrete), and two plausible visual representations each. The examples are extracted from the \textit{Bing} dataset described in Section~\ref{subsec:Images}.}
	\label{diversity}
\end{figure*}

Various studies have successfully attempted to predict the concreteness score of a concept by exploiting the visual information extracted from multiple images associated with it in combination with more traditional textual representations \cite{KielaEtAl:14, HesselEtAl:18, charbonnier-wartena-2019-predicting}. A building assumption of these visual models is a certain degree of visual coherence that facilitates the construction of stable visual representations. 
While images of concrete concepts are generally expected to show greater consistency, a notable variability is still present in \textit{both} concrete and abstract concepts, 
i.e., the properties of these images, including color, shape, size, and other visual details, may vary significantly, thus reflecting the diversity of the intrinsic nature of the concept. 

Figure~\ref{diversity} illustrates the variability in the images associated with abstract and concrete concepts. Images can be highly representative of a concept and be visually similar (e.g., \texttt{affordability, waterfall}) or be rather different from one another (e.g., \texttt{chariot}). Conversely, images of a concept can be very similar but not informative representations of the concept (e.g., \texttt{allegiance}). Finally, they can be highly different yet individually all strongly associated with the same target concept (e.g., \texttt{banana, accountability}). These degrees of variation highlight some of the inherent challenges computational methods face in constructing a comprehensive visual representation of a concept and mapping it to its labels. These challenges are orthogonal to previously raised issues regarding depictions of (mostly concrete) semantic concepts such as variability of prototypicality \citep{GualdoniEtAl:23,HarrisonEtAl:23,TagliaferriEtAl:23}, and will be explored further in the course of this study (see RQ3 below).

Our research targets the challenges of precisely quantifying the contribution of visual information in describing concrete versus abstract concepts, using interpretable representations to explore the following three research questions:

\noindent
\textbf{RQ1}: Can visual diversity differentiate between concrete and abstract concepts?

\noindent
\textbf{RQ2}: How consistent are visual attributes across multiple images of the same concept? 

\noindent
\textbf{RQ3}: What are inherent yet plausible failure categories for unimodal visual representations?

In Study~1, we address \textbf{RQ1} by classifying approximately $500$ concrete and $500$ abstract concepts based on the diversity in visual features extracted from images associated to each concept.
This approach helps us identify the most salient visual features that distinguish between concepts based on their concreteness. 
In Study~2, we address \textbf{RQ2} and analyse the consistency of these features across multiple concept images, by performing a nearest-neighbor analysis of image representations.
Finally, Study~3 targets \textbf{RQ3} by qualitatively analyzing the failures in Study~2 and manually determining categories of problematic issues.

To our knowledge, this is the first large-scale study conducting a detailed quantitative and qualitative investigation into how visual features
contribute to representing abstract and concrete concepts. By focusing exclusively on the visual component, we can systematically identify the strengths and weaknesses of using such extremely rich source of information. Additionally, compared to previous studies, our methodology highlights cases that are particularly challenging because they are equally plausible rather than erroneous.

\section{Related Work}
\label{sec:related}

The distinction between abstract and concrete words is highly relevant for natural language processing and has been exploited for metaphor detection \citep{TurneyEtAl:11, TsvetkovEtAl:13, Koeper/SchulteImWalde:16b, MaudslayEtAl:20, su2021multimodal, Piccirilli/SchulteImWalde:22a},
lexicography \citep{Kwong:11},
and embodied agents and robots \citep{Cangelosi/Stramandinoli:18, RasheedEtAl:18, AhnEtAl:22}, among others. 
Most studies addressing this distinction have primarily focused on the textual modality alone
\citep{FrassinelliEtAl:17b, LjubesicEtAl:18, NaumannEtAl:18, charbonnier-wartena-2019-predicting, Frassinelli/SchulteImWalde:19, SchulteImWalde/Frassinelli:22, TaterEtAl:22}. First extensions to further modalities explored free associations and imageability \citep{HillEtAl:14, KielaEtAl:14, Koeper/SchulteImWalde:16b}. Since the primary distinction between degrees of abstractness is influenced by the strength of sense perception, with vision being considered the main source of perceptual information \citep{BrysbaertEtAl:14, LynottEtAl:20}, later studies began to explore bimodal approaches that combine text and images \citep{BhaskarEtAl:17, HesselEtAl:18}. 
Compared to text, the visual component has provided less definitive insights, and it is unclear whether this is due to architectural choices or to the inherent challenge triggered by depicting abstract concepts. \citet{cerini-etal-2022-speed} analyzed the mechanism behind this indirect grounding of abstract concepts by collecting word association data and pairs of images and abstract words. \citet{kastner2019estimating} discuss the visual variety of a dataset using mean shift clustering, where the dataset is designed to contain images in the same ratio of sub-concepts as in real life. \citet{kastner2020estimating} performed a regression study to predict the imageability of concepts using the YFCC100M dataset; our feature selection builds on their results. \citet{KielaEtAl:14} and \citet{HesselEtAl:18} postulated that concreteness in images varies across datasets and is not directly connected to the underlying linguistic concept. \citet{pezzelle2021word} evaluated the alignment of semantic representations learned by multimodal transformers with human semantic intuitions, finding that multimodal representations have advantages with concrete word pairs but not with abstract ones. \citet{Vaze_2023_CVPR} argue that there are multiple notions of ``image similarity'' and that models should adapt dynamically. For example, models trained on ImageNet tend to prioritize object categories, while a user might want the model to focus on colors, textures, or specific elements in the scene. They introduce the GeneCIS benchmark, assessing models' adaptability to various similarity conditions in a zero-shot evaluation setting. They observe that even robust CLIP models struggle to perform well, and performance is only loosely connected to ImageNet accuracy. Most recently, \citet{tater-etal-2024-evaluating} examined to which degree SigLIP, a state-of-the-art Vision-Language model (VLM), predicts labels for images of abstract and concrete concepts that are semantically related to the original labels in various ways: synonyms, hypernyms, co-hyponyms, and associated words. The results show that not only abstract but also concrete concepts exhibited significant variability in semantically appropriate label variants.

%%%%%%%%%%%%%%%%%%%%%%%%%%%%%%%%%%%%%%%%%%%%%%%%%%%%%%%%%%%%%%%%%%%%%%%%%%%%%%%%%%%%%%

\section{Experimental Design}
\label{sec:experiments}
In the following sections, we present the resources used in our analyses. We introduce the target concepts under investigation, their abstractness scores, and the associated images. Subsequently, we describe the algorithms employed to extract the visual attributes from the images.

\subsection{Target Concepts \& Concreteness Norms}
To select a balanced amount of concrete and abstract targets, we use the concreteness ratings from \citet{BrysbaertEtAl:14} (henceforth, \textit{Brysbaert norms}) that were collected via crowd-sourcing, and range from $1$ (clearly abstract) to $5$ (clearly concrete). Our analyses focus on $500$ highly abstract (concreteness range: $1.07-1.96$) and $500$ highly concrete ($4.85-5.00$) nouns. We excluded nouns with mid-range concreteness scores as they are typically more challenging for humans and thus lead to noisier distributional representations \citep{Reilly/Desai:17, Pollock:18, KnuplesEtAl:23}. 

\subsection{Image Datasets}
\label{subsec:Images}

We extracted images for each target noun -- both concrete and abstract -- from two distinct datasets: (i) the YFCC100M Multimedia Commons Dataset (\textit{YFCC};  \citet{yfcc}); and (ii) \textit{Bing}\footnote{\url{https://www.bing.com/images/}}. 
 
For the YFCC dataset, we randomly selected $500$ images tagged with each target concept. The YFCC dataset is the largest publicly available user-tagged dataset containing $100$ million media objects extracted from the online platform Flickr. Its images exhibit diversity in quality, content, visual coherence, and annotation consistency. Thus, we use them to test the robustness of the methods adopted and support the ecological validity of our studies despite introducing a significant level of noise from variable image quality and annotation inaccuracies.

For the Bing dataset, the images were selected by directly querying the target word. 
To avoid duplicates, we automatically excluded images where all the pixel values were exactly the same as another image and downloaded new ones if necessary (continuing recursively).
Subsequently, we manually inspected the remaining images for  inappropriate content (e.g., sexual content) and removed them. We kept a maximum of $25$ images for each target concept as this was the highest number consistently available across all target concepts. Given that Bing was our control condition, maintaining a balanced dataset was important. 
Finally, for both YFCC and Bing, we only included images with a size of $256\times256$ pixels or higher and resized them to a uniform size as required for each feature analysis. 

Despite the huge size of the YFCC dataset, we were unable to extract the desired number of $500$ images across all our $1,000$ targets ($500$ concrete and $500$ abstract). Table~\ref{tab:numImagesLost} shows for how many concrete and abstract target nouns we were able to retrieve $25 \ldots 500$ images. For example, we could only retrieve $500$ images for subsets of $463$ concrete and $151$ abstract nouns. For the following analyses, it is therefore important to remember that abstract targets are more affected than concrete targets regarding the available numbers of images.

\begin{table}[!tbh]
\centering
\footnotesize

\begin{tabular}{@{}crrrrrr@{}}
\toprule
 \textbf{ $\#$ Images} & \textbf{25} & \textbf{100} & \textbf{200} & \textbf{300} & \textbf{400} & \textbf{500} \\ \\ \midrule

\textbf{Concrete} & 498 & 494 & 481 & 475 & 472 & 463 \\
\textbf{Abstract} & 420 & 304 & 237 & 197 & 172 & 151 \\

\bottomrule
\end{tabular}
\caption{Number of abstract and concrete target nouns for different number of images per target (YFCC).}
\label{tab:numImagesLost}
\end{table}

\subsection{Extraction of Visual Attributes}
\label{sec:features}
When evaluating an image, it is crucial to consider the visual properties that help us capture its most prominent characteristics. We extracted a series of 
independent visual features (attributes) for each image associated with our target words. Furthermore, we utilized two SOTA visual models to generate comprehensive image representations and use them as benchmarks for our analyses. 

We start with low-level features, including colors, shapes, and textures. Colors are described as distributions in the \textit{HSV} space: hue, saturation, value \citep{joblove1978color}. Shapes and structures in an image are quantified using the \textit{Histogram of Oriented Gradients} (\textit{HOG}; \citet{dalal2005histograms}): this feature descriptor captures the occurrences of gradient orientation in localized image segments. We capture texture information using two methods: the \textit{Gray-Level Co-occurrence Matrix} (\textit{GLCM}; \citet{haralick1973GLCM}) and the \textit{Local Binary Patterns Histograms} (\textit{LBPH}; \citet{ojala2002LBPH}). GLCM is a statistical measure that considers the spatial relationship of pixels represented as a co-occurrence matrix. This approach quantifies how often pairs of pixel values appear together at a specified spatial orientation.
LBPH, on the other hand, calculates a local representation of texture by comparing each pixel with its neighbors. 

We also include more complex features representing objects and their relationships in a scene. Low-dimensional abstract representations of a scene are computed using \textit{GIST} \cite{oliva2001GIST}. To identify similar sub-regions and patches across images, we use the Speeded-Up Robust-Features feature descriptor (\textit{SURF}; \citet{bay2008speeded}) combined with a Bag-of-Words model (\textit{BOW}; \citet{csurka2004BOW})  using k-means clustering. The objects occurring in an image are detected using the YOLO9000 model (\textit{YOLO}; \citet{redmon2017yolo9000}) pre-trained on $9,418$ object classes. We then extract hypernymy relationships from WordNet \cite{miller1995wordnet} to reduce the number of object types detected from the original $9,418$ to $1,401$ classes of hypernyms.  
With this approach, we substantially alleviate sparsity while retaining most of the information captured by the model since the hypernyms contain information specific enough to qualify the objects in an image. We then determine the location of the objects detected in the image and quantify their spacial relationship by using an overlapping $10\times10$ grid and counting the number of objects co-occurring in each cell. On average, only $10\%$ of the images associated with each target noun contain an object detected by the YOLO model, even though $330$ of our $500$ concrete concepts are also in the $9,000$  object classes in YOLO (for more details, see Table~\ref{stats_yolo_available} in the Appendix). 
 
Finally, we generate comprehensive visual representations with two pre-trained models for feature extraction: \textit{SimClr} \cite{chen2020simple} and \textit{Vision Transformer} (\textit{ViT}; \citet{dosovitskiy2020VIT}). We use these models as a benchmark against basic features since they are more advanced models and are the backbone of most currently used multi-modal models (e.g., CLIP uses a ViT encoder). SimClr builds image representations using contrastive learning trained on images only. It maximizes the agreement between differently augmented views of the same image using a contrastive loss. 
ViT is a supervised model for image classification trained by splitting an image into patches, which are then combined and converted into linear embeddings using a transformer network. ViT uses attention maps to deduce an image's most informative parts. It is pre-trained on the ILSVRC-2012 ImageNet dataset with $1,000$ classes. Only $36$ of our target concepts completely overlap with these $1,000$ classes, indicating that our results are generalizable and not the consequence of the overlap between the classes from ImageNet and our target concepts.

\vspace{+2mm}
\subsubsection{Feature Combination}
\label{subsub:FeatCombination}
As traditionally done in the literature (e.g., \citet{KielaEtAl:14, BhaskarEtAl:17}), we create one single visual representation for each concept combining the information from the different images. To achieve this, we compare the feature vectors of all images of the same concept. This results in nine square similarity matrices (one per visual attribute) of size $N \times N$ (the number of images), which are symmetrical. These matrices capture the characteristics of a concept and, at the same time, highlight the variability across its different visual representations.
Given that the similarity matrix's values depend on the order of the images, we calculate the $N$ eigenvalues of each similarity matrix to provide an invariant representation that is order-independent. 
This also helps us reduce the dimensionality of features and make them consistent, while still encoding the core characteristics of each feature.

%%%%%%%%%%%%%%%%%%%%%%%%%%%%%%%%%%%%%%%%%%%%%%%%%%%%%%%%%%%%%%%%%%%%%%%%%%%%%%%%%%%%%%

\vspace{+2mm}
\section{Study 1: Classifying Concepts using Visual Information}
\label{sec:study1}
This first study aims to identify the visual features that are most useful 
for discriminating between images of concrete vs.\ abstract nouns.
We utilize three different classifiers: Support Vector Machine (SVM) with \textit{rbf} kernel, Random Forest (RF), and Logistic Regression (LR) with hyper-parameter tuning, while using the eigenvalues of the combined visual features described above as predictors.\footnote{We also conduct a regression analysis and present the results in the Appendix.} In the main text, we report the performance of the RF model as, overall, it outperforms the other two classifiers (the results for LR and SVM are reported in Figures~\ref{fig:classification_ALL_LR} and~\ref{fig:classification_ALL_SVM} in the Appendix). We evaluate the predictive power of our features independently and by concatenating them.  
To account for data skewness between classes, we apply $5$-fold cross-validation.

\subsection{Results}
\begin{figure}[!thb]
     \includegraphics[scale=0.4]{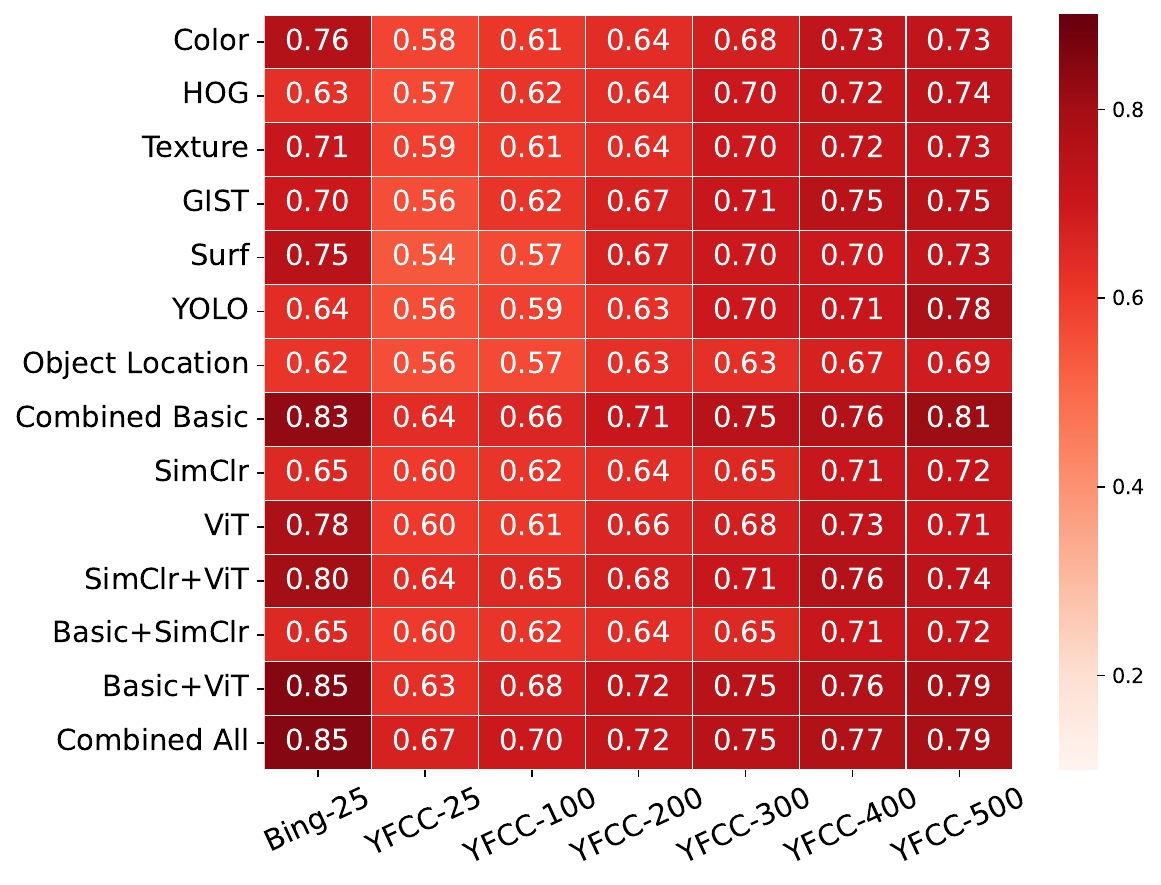}
     
 	\caption{Weighted F1-scores for different features and different dataset sizes for Bing and YFCC using RF.}
 	\label{classification_combined}
 \end{figure}

\begin{figure*}[!thb]
 	\center
     \includegraphics[width=1\textwidth]{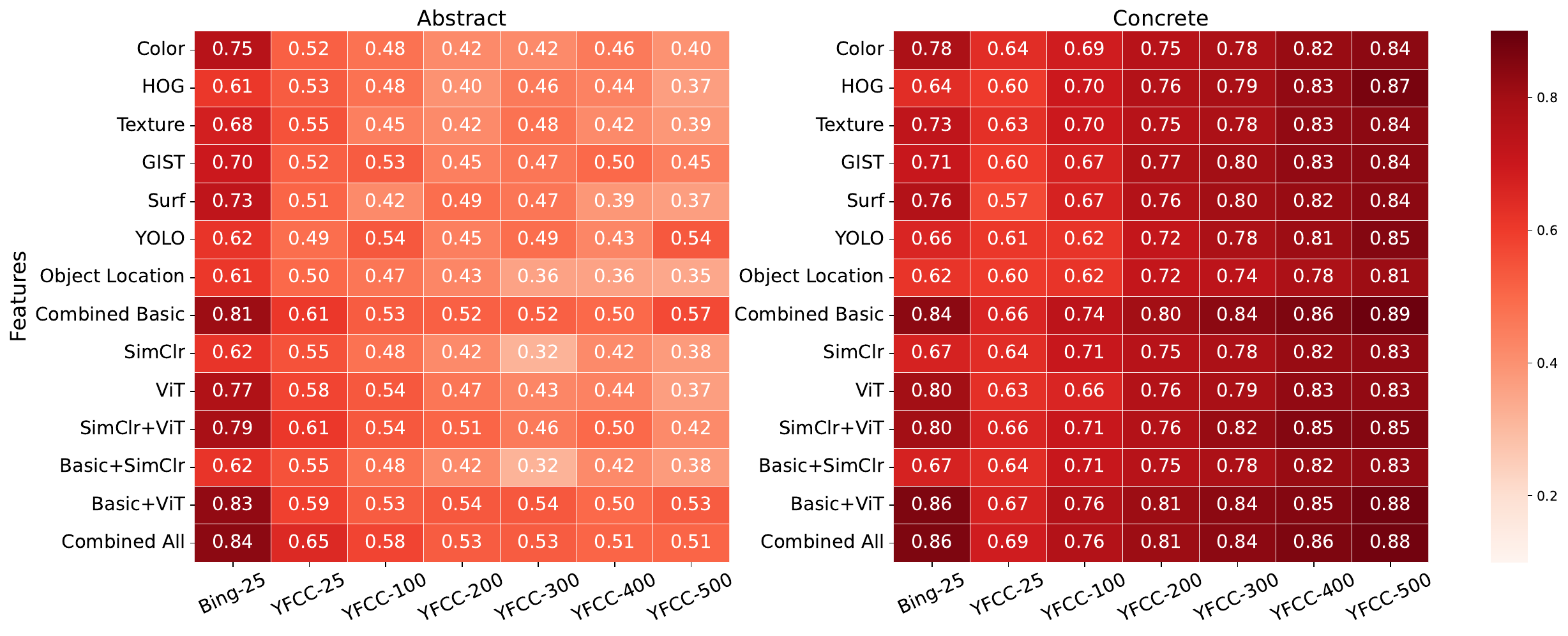}
     
 	\caption{Class-wise F1-scores of abstract and concrete concepts for RF, across features and dataset sizes.}
      \vspace{+2mm}
 	\label{classification_classwise}
 \end{figure*}

Figure~\ref{classification_combined} reports the F1-scores obtained by the RF classifier. We compare the performance of low-level visual features used individually and in combination (Combined Basic), as well as advanced features derived from ViT and SimClr, along with their combinations. The different columns reflect the number of images available for each target. Notably, the model trained on a mix of only basic features consistently obtains the highest F1-scores (darker color) across all datasets and image counts. Incorporating more sophisticated visual features, such as SimClr or ViT, offers limited advantages and only when merged with the basic feature set. 
When comparing the performance for Bing and YFCC, images extracted from Bing consistently outperform those from YFCC across all feature types and number of images. Furthermore, a trend emerges with YFCC images: increasing the number of images from $25$ to $500$ leads to a steady improvement in performance.

In Figure~\ref{classification_classwise} we report the same results but separately for abstract vs.\ concrete concepts. It is striking to see that, on average, the RF model classifies more effectively concrete than abstract concepts, simply based on their visual diversity. We also see that while adding visual information is beneficial for classifying concrete nouns, it is detrimental for abstract nouns. This is strongly influenced by the marked reduction in the abstract target nouns when increasing the number of images (see Table~\ref{tab:numImagesLost}).

\subsection{Discussion}

This study tested how reliable are visual attributes in capturing the diversity of images to distinguish between concrete vs.\ abstract concepts. Overall, low-level features like color and patch similarity (SURF) play a more vital role in predicting abstractness than more complex feature types like object location and detection. This suggests that while high-level object information may vary considerably, low-level features remain more consistent across different depictions of the same concept, which is crucial in classifying concepts based on their abstractness. 
This observation extends to more sophisticated feature representations such as ViT and SimClr as well. 

The results in Figure~\ref{classification_classwise} show that for coherent and less noisy images of concepts in the Bing dataset, the model shows comparable performance for both concrete and abstract nouns, mirroring the general patterns discussed above. However, when increasing the number of images for the YFCC dataset, the performance of the model progressively increases for concrete nouns with the addition of more images while the performance for abstract nouns decreases. Particularly when evaluating the performance of the model with $500$ images per concept, it becomes evident that basic features are all very good predictors (all above 0.84) of concreteness. Notably, also more complex features, such as YOLO and object location, show a steady improvement and achieve a level of performance that closely aligns with that of the simpler, low-level features regardless of the low number of objects detected.  Once again, the use of more sophisticated representations does not show any substantial improvement in the performance of the models. When examining abstract nouns, the drastic reduction in the number of target nouns with the addition of more images inevitably impacts the performance of the model in a negative way. This reduction renders any subsequent analysis of this particular subset less informative.

\vspace{+2mm}
\section{Study 2: Inspecting Visual Nearest Neighbors}
\label{sec:study2}
In our second study, we directly build on the evidence from Study~1 and perform a nearest neighbors analysis to inspect the consistency of visual attributes across multiple images of the same concept.
We compute the cosine similarity of each image of a concept with all other images of all concepts in the same dataset, represented by using the same features as before.
We then inspect the top $N$ (where $N = [25, \dots , 500]$) most similar images and compute the percentage of neighbors associated with the same concept; e.g., how many nearest neighbor images of an image of \texttt{banana} are also images of \texttt{banana}. 

\subsection{Results}
\label{subsec:results2}

\begin{figure*}[!thb]

 	\center
     \includegraphics[width=0.95         \textwidth]{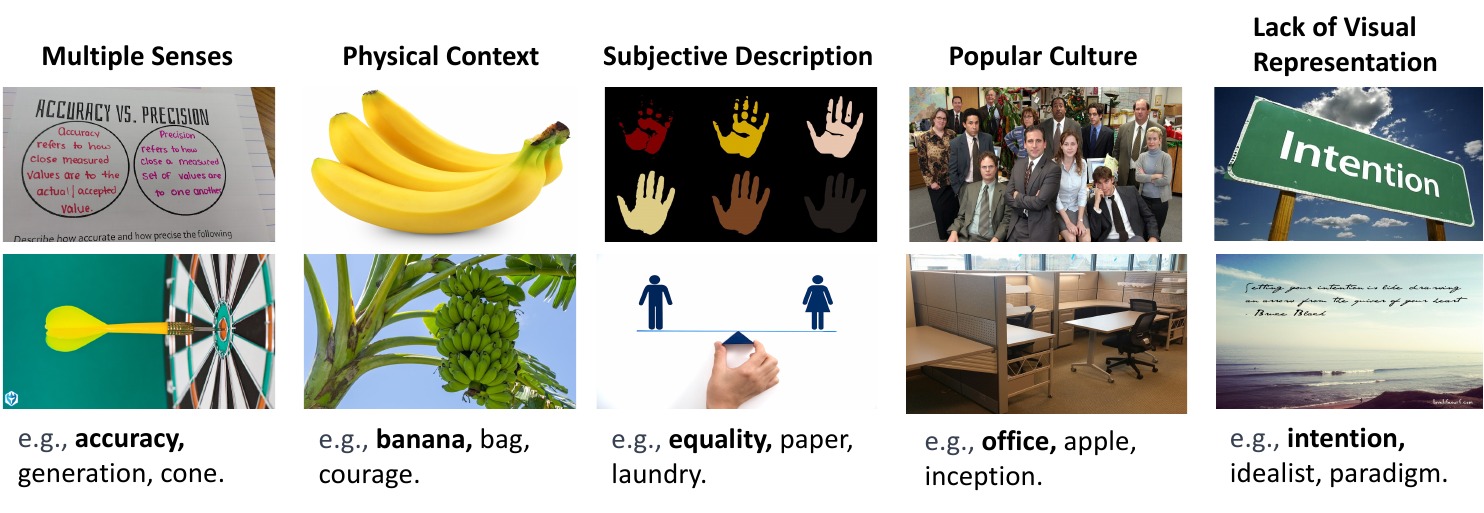}
     
 	\caption{Five most frequent reasons (top row) of visual diversity among images associated with the same concept (indicated by the bold font in the example list below each image).}
  \vspace{+1mm}
 	\label{fig:challenges}
 \end{figure*}

\vspace{+2mm}
Table~\ref{tab:results_nn_combined} presents the average percentage of visual neighbors associated with the same concept across different features for the Bing, YFCC-25 and YFCC-500 datasets (see Table~\ref{tab:results_nn_variable_images} in the Appendix for YFCC-100, 200, 300 and 400). 
Overall, the results across features and datasets are very low. On average, less than $1\%$ of the images closest to a specific target are associated with it, both for concrete and abstract targets, but interestingly exhibiting divergent patterns. With Bing, even though not as strongly as we initially expected, the nearest neighbors of concrete concepts show a higher similarity than those of abstract concepts. Among simple features, object detection (YOLO) marginally outperforms the rest. However, for the YFCC-25 dataset, all basic features except object location produce better results for abstract concepts. When we include more images (YFCC-500), the percentage of correct neighbors drop even more. Unlike the results of the classification study discussed in Section~\ref{sec:study1}, employing more sophisticated representations, such as Vision Transformer, yields the best outcomes, although the performance levels remain low. Moreover, abstract concepts in the YFCC-25 dataset perform similarly to, or even better than, their counterparts in the Bing dataset, despite still showing overall poor performance.

\begin{table}[!tbh]
\centering
\footnotesize
\begin{tabular}{@{}lrrrrrr@{}}
\toprule
 & 
  \multicolumn{2}{c|}{ \textbf{Bing-25}} &
  \multicolumn{2}{c|}{ \textbf{YFCC-25}} 
  &
  \multicolumn{2}{c}{  \textbf{YFCC-500}}  \\ \cmidrule(lr){2-7} 
\multirow{-2}{*}{\textbf{Attribute}} &
  \multicolumn{1}{c}{ \textbf{A}} &
   \multicolumn{1}{c}{\textbf{C}} &
  \multicolumn{1}{c}{ \textbf{A}} &
   \multicolumn{1}{c}{\textbf{C}} &
  \multicolumn{1}{c}{ \textbf{A}} &
   \multicolumn{1}{c}{\textbf{C}} \\ \midrule
Color& 
$0.68$ & $0.96$ & $1.70$ & $0.95$ & $0.81$ & $0.65$ \\ 
HOG         & $0.48$ & $1.44$ & $0.68$ & $0.58$ & $0.36$ & $0.44$ \\ 
Texture     & $0.29$ & $0.33$ & $0.35$ & $0.26$ & $0.28$ & $0.27$ \\ 
GIST        & $0.55$ & $1.88$ & $1.03$ & $0.76$ & $0.52$ & $0.56$ \\ 
SURF        & $0.64$ & $1.70$ & $0.93$ & $0.62$ & $0.40$ &$0.38$ \\ 
YOLO        & $2.25$ & $3.19$ & $1.09$ & $1.03$ & $1.64$ & $1.57$ \\ 
Object Loc. & $0.18$ & $0.39$ & $0.15$ & $0.18$ & $0.24$ &$0.27$ \\ \midrule
Combined    & $0.64$ & $2.14$ & $1.40$ & $0.99$ & $0.69$ & $0.75$ \\ \midrule
Simclr      & $0.65$ & $1.49$ & $1.15$ & $0.79$ & $0.53$ &$0.55$ \\ 
ViT         & $2.83$ & $26.44$ & $3.71$ & $6.67$ & $2.27$ & $6.63$ \\ 
\bottomrule
\end{tabular}
\caption{Average percentage of nearest neighbors (out of top $25$ or $500$, respectively) associated with the same abstract (A) or concrete (C) concept.} 

\label{tab:results_nn_combined}
\end{table}
\subsection{Discussion}
\label{subsec:discussion2}

This study demonstrated that images with similar labels share very little visual information. While \citet{HesselEtAl:18} and \citet{hewitt-etal-2018-learning} have already discussed the lack of a univocal visual representation for abstract concepts, our results reveal a more nuanced pattern. Surprisingly, we found significant visual variability even among concrete concepts, which challenges the assumption that images of the same target share consistent visual features. 
More complex models (like ViT) can capture the higher agreement between concrete concepts, indicating that images of concrete concepts are generally more consistent or similar. However, basic features may encode more distinctive information related to individual abstract concepts than concrete concepts. Moreover, combined basic features, which performed better than ViT in Study~1, do not encode enough information for nearest neighbors compared to ViT.

\vspace{+1mm}
\section{Study 3: Exploring Factors Behind Visual Diversity}

\label{sec:study3}

As discussed in Section~\ref{sec:study2} when analyzing nearest neighbors, the biggest challenge in using images of a concept comes from the diversity of the images associated with it. The same concept, whether abstract or concrete, can be depicted in many different yet plausible ways, thus relating to previously discussed issues regarding the variability of prototypical attributes in depictions of the semantic concepts 
\citep{GualdoniEtAl:23,HarrisonEtAl:23,TagliaferriEtAl:23}. In our final analysis, we provide a manual classification of the critical factors influencing the nearest neighbors of our target concepts.

We identified five primary reasons for visual diversity, as exemplified in Figure~\ref{fig:challenges}. For concepts like \texttt{accuracy}, \texttt{generation}, and \texttt{cone}, the words used as a proxy to our concepts may be lexically ambiguous and have \textit{multiple senses}. According to Wordnet \cite{miller1995wordnet}, $650$ out of the $918$ concepts used in our studies have more than one sense, and $248$ concepts have four or more senses. A further source of variability is \textit{physical context}, manifesting itself as different background information, objects, etc. In our example, both images depict \texttt{bananas}, but they differ visually: in the bottom image, the bananas are still hanging on a banana tree, which dominates the scene. Another form of visual diversity is triggered by \textit{subjective representations}: concepts like \texttt{equality} and \texttt{paper} show very high variability. People have different visual interpretations and realizations of these concepts, even when the underlying conceptual meaning is understood in the same way. \textit{Popular culture} often associated with films and books represents a kind of variability that introduces visual representations often completely disjoint from the original meaning of the concept: for example, images of the concepts \texttt{inception} and \texttt{office} contain images which are from the movie ``Inception'' and the TV show ``The Office'', respectively. Finally, primarily abstract concepts like \texttt{intention} and \texttt{idealist} \textit{lack distinctive visual representations} and are hard to depict. These concepts are often represented by writing the associated word into an image. 

Another orthogonal source of variability in visual representations comes from the selection process in the source dataset. For example, the YFCC dataset contains images from Flickr that are uploaded by users, resulting in a lot of variability and bias toward specific senses of a tagged concept. 

\subsection{Experimental Design and Results}
Given that we are the first suggesting this categorization of "challenges" related to very diverse but still plausible images associated with specific concepts, we ask $13$ participants to evaluate our five categories using a subset of target images related to abstract and concrete concepts. We selected two images each for a subset of $18$ concepts, while ensuring that we included potentially ``problematic" cases. The experiment was conducted on Google Forms, where the participants could choose at least one reason (and possibly more) why two images of the same concept differed.

Figure~\ref{imsstudy} in the Appendix presents the $18$ concepts, the image pairs, and the results of the annotation. 
For most of the target images, we see high agreement between annotators on a specific ``reason for visual diversity'', with Krippendorff’s~$\alpha=0.29$ \cite{krippendorff80, Artstein/Poesio:08}. For example, $12$ out of $13$ ratings assigned the visual ambiguity for the concept \texttt{banana} to variability in the \textit{physical context}. And $10$ out of $16$ ratings for \texttt{intention} are linked to a \textit{lack of visual representation}. To further inspect the variability and complexity of plausible, but yet diverse, visual representations across images of these $18$ images, we set up an Amazon Mechanical Turk\footnote{\url{https://www.mturk.com}} study where nine native English speakers (from the UK and USA) had to describe in one word "what is depicted in an image". As an example of the plausible variability in the response, when evaluating the response for the image of \texttt{equality} showing six colorful hands (see Figure~\ref{fig:challenges}), $27$ out of $39$ participants listed words referring to the colors in the image. Even though colors provide relevant attributes of the image, they do not represent generally salient meaning components of the associated concept. See Tables~\ref{tab:amt_annotations_abs} and \ref{tab:amt_annotations_conc} in the Appendix for the complete lists of words generated for the $10$ images associated with concrete and abstract concepts.\footnote{The complete dataset of human-generated words (manually checked for offensive content) can be found here: \\
\url{https://github.com/TarunTater/AbstractConceptsInImages/tree/main/depict_image_annotations}}

\vspace{+2mm}
\section{Conclusion}
\label{sec:conclusion}
We performed three empirical studies to understand how abstract and concrete concepts are depicted in images. Compared to existing studies, we focused exclusively on the role of variability in the visual information. After automatically generating nine different feature representations for the images, we tested their reliability in a classification study to distinguish between concrete and abstract concepts. We showed that, overall, combining low-level features produces good results. We then investigated the consistency of the visual attributes across multiple images of the same concept by looking at the nearest neighbors of each image in the two datasets. The results across feature types, datasets, and concreteness scores were very low; overall, abstract concepts showed considerably higher cases where none of the most similar images were associated with the same concept. The results also showed that both concrete and abstract concepts lack a univocal visual representation in terms of objects depicted and, in general, basic visual properties. Finally, in an error analysis study with human participants, we highlighted the five most frequent reasons explaining visual diversity among images associated with the same concept. 

Overall, our research significantly advances the understanding of the role of the visual component in tasks that heavily rely on the integration of multiple types of information beyond just text.

\section*{Limitations}
The number, random selection, and content of the images used in this study may introduce some variability in the results. Moreover, any interpretation based on the output of the object detection systems should be made with caution, especially considering the very low number of images where an object was detected.

\section*{Ethics Statement}
We see no ethical issues related to this work. All experiments involving human participants were voluntary, with fair compensation (12 Euros per hour), and participants were fully informed about data usage. We did not collect any information that can link the participants to the data.  
All modeling experiments were conducted using open-source libraries, which received proper citations.

\section*{Acknowledgements}

This research is supported by the DFG Research Grant SCHU 2580/4-1 \textit{Multimodal Dimensions and Computational Applications of Abstractness}. We thank Allison Keith, Amelie Wührl, Ana Baric, Christopher Jenkins, Filip Miletic, Hongyu Chen, Iman Jundi, Lucas Moeller, Maximilian Martin Maurer, Mohammed Abdul Khaliq, Neele Falk, Prisca Piccirilli, Simon Tannert, Tanise Ceron, and Yarik Menchaca Resendiz for their help in the evaluation tasks.

\bibliography{ssiw,srw}

\newpage
\clearpage

\section{Appendix}

\subsection{Feature Availability - YOLO}
As mentioned in Section~\ref{sec:features}, there are instances where the YOLO9000 model does not detect any objects in an image. In Table~\ref{stats_yolo_available} we examine the percentage of images per concept where at least one object was detected. On average, around $10$\% of the images associated to an abstract concept have at least one  object detected. For concrete concepts, this value is slightly lower, ranging between $8.5\%$ to $9.5$\%. We hypothesize that the surprisingly low number of images where objects are detected (only $10\%$ of the images) is very likely due to the following reasons. Firstly, the YFCC dataset exhibits high visual variability in terms of informativeness and quality of the user-tags used on Flickr. For example, a tag like 'dessert' might be attributed to vastly different types of images, ranging from cakes to fruit platters or ice creams. In such cases, the user tag may describe concepts or objects that fall under the same broad category but differ from the specific items the object detection model is trained to recognize. Some tags may also refer to objects that are not very salient in the visual scene, making them difficult for the model to detect. This mismatch between the user tags and the model's ability to identify objects likely contributes to the low detection rate observed. Moreover, we used YOLO9000 (released in $2016$--$17$) because it is the only available model with $9,000$ classes, even though there are more powerful object detection models (YOLOv9) available. For our task, this was one of the crucial reasons for selecting the model. We wanted to detect as many object classes as possible since we can not know which of these object classes may be present within images of concepts, especially for abstract concepts.

\begin{table}[!tbh]
\centering
\footnotesize

\begin{tabular}{@{}lrr@{}}
\toprule

 & 
\multicolumn{2}{c}{\textbf{Number of Images ($\%$)}}  \\
\textbf{Dataset}  & 
\multicolumn{1}{c}{\textbf{A}} & \multicolumn{1}{r}{\textbf{C}} \\ \midrule
YFCC - $500$ & $10.02$ & $9.48$ \\

YFCC - $400$ & $10.01$ & $9.37$ \\

YFCC - $300$ & $10.07$ & $9.28$ \\

YFCC - $200$ & $10.09$ & $9.06$ \\

YFCC - $100$ & $10.00$ & $8.87$ \\

YFCC - $25$ & $10.08$ & $8.64$ \\

Bing - $25$ & $14.28$ & $15.28$\\
\bottomrule
\end{tabular}
\caption{Average number (percentage) of images for abstract (A) and concrete (C) concepts containing at least one object detected by the YOLO9000 model.}  
\label{stats_yolo_available}
\end{table}

\subsection{Classification Results for Different Classifiers}

\begin{figure*}[!th]
 	\center
      \includegraphics[width=0.95\textwidth, height=5cm]{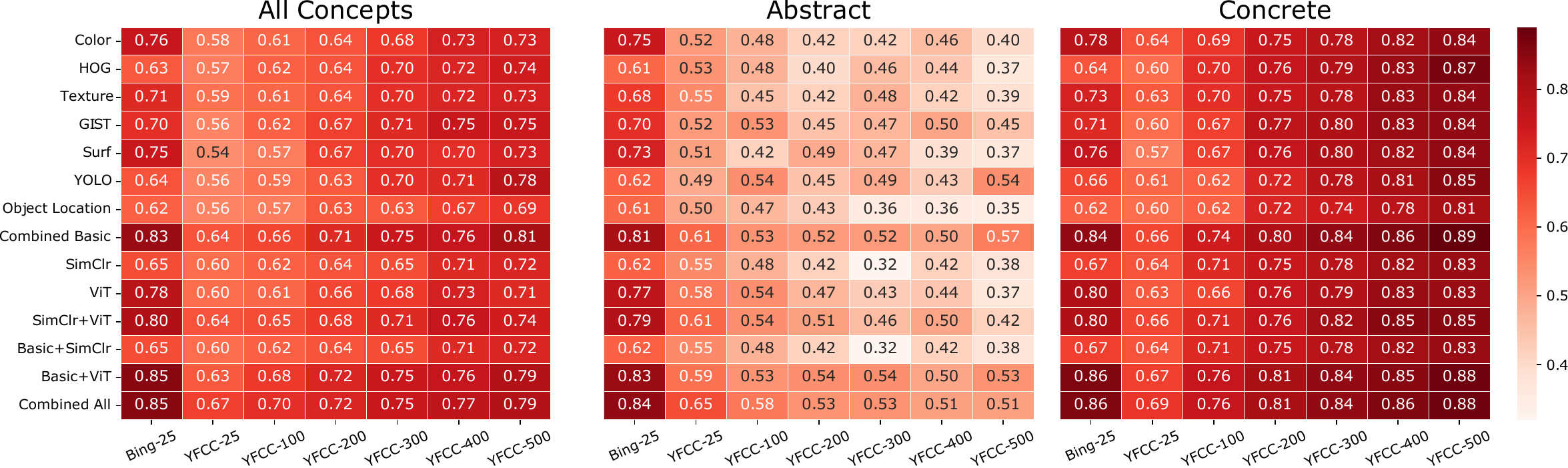}
  	\caption{Weighted F1-scores (overall and by class) for different features and different dataset sizes for Bing and YFCC using \textbf{Logistic Regression}.}
 	\label{fig:classification_ALL_LR}
 \end{figure*}

\begin{figure*}[!th]
 	\center
      \includegraphics[width=0.95\textwidth, height=5cm]{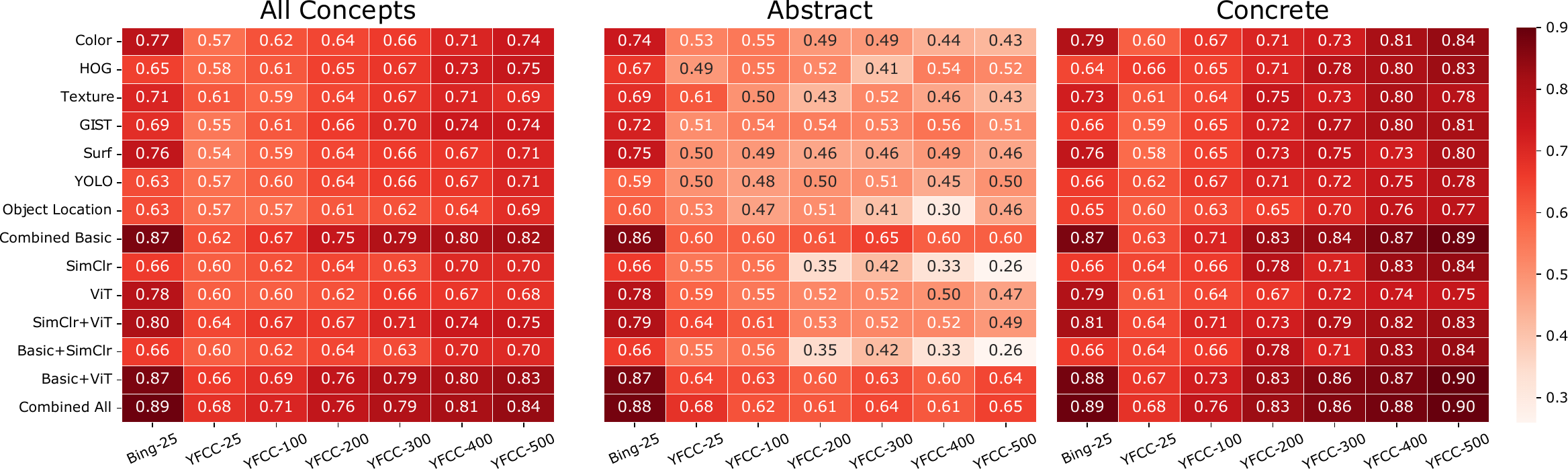}
 	\caption{Weighted F1-scores (overall and by class) for different features and different dataset sizes for Bing and YFCC using \textbf{Support Vector Machines}.}
 	\label{fig:classification_ALL_SVM}
 \end{figure*}

In the classification study in Section~\ref{sec:study1}, we experimented with three different classifiers: Support Vector Machines (SVM) with \textit{rbf} kernel, Random Forests (RF), and Logistic Regression (LR). The results for the RF model are reported in the main text (Figures~\ref{classification_combined} and ~\ref{classification_classwise}). The results, combined and by class, for Logistic Regression can be found in Figure~\ref{fig:classification_ALL_LR}. The results for SVM are presented in Figure~\ref{fig:classification_ALL_SVM}.

\subsection{Eigenvalues and How to Infer Them?}
We use eigenvalues to extract characteristics of the similarity matrix in Study $1$. 
The top eigenvalues capture the most information about the similarity matrix as they represent the variance of principal components. Hence, they are expected to have the most information on the similarity/variance of images. So, all high eigenvalues would indicate very diverse images for a feature, whereas all low eigenvalues would suggest high similarity. 

\subsection{Nearest Neighbor Results}
Table~\ref{tab:results_nn_variable_images} supplements the results shown in Table~\ref{tab:results_nn_combined} by incorporating the nearest neighbor analysis with varying quantities of images per concept (ranging from 100 to 400) extracted from the YFCC dataset. 

\subsection{Cosine Similarity Comparison between Abstract and Concrete Concepts}

\begin{table}[!tbh]
\centering
\footnotesize
\begin{tabular}{@{}lrrrrrr@{}}
\toprule
 &
  \multicolumn{2}{c|}{ \textbf{Bing-25}} &
  \multicolumn{2}{c}{  \textbf{YFCC-25}}  \\ \cmidrule(lr){2-5}
\multirow{-2}{*}{\textbf{Attribute}} &
  \multicolumn{1}{c}{ \textbf{A}} &
   \multicolumn{1}{c}{\textbf{C}} &
  \multicolumn{1}{c}{ \textbf{A}} &
   \multicolumn{1}{c}{\textbf{C}} \\ \midrule

Color       & $0.91$ & $0.92$ & $0.92$ & $0.92$ \\ 
HOG         & $0.78$ & $0.80$ & $0.80$ & $0.81$ \\
Texture     & $0.99$ & $0.99$ & $0.99$ & $0.99$ \\
GIST        & $0.91$ & $0.91$ & $0.93$ & $0.93$ \\
SURF        & $0.61$ & $0.64$ & $0.42$ & $0.42$ \\
YOLO        & $0.95$ & $0.89$ & $0.91$ & $0.86$ \\
Object Loc. & $0.85$ & $0.84$ & $0.81$ & $0.80$ \\ \midrule
Combined    & $0.98$ & $0.98$ & $0.98$ & $0.98$ \\ \midrule
Simclr      & $0.98$ & $0.98$ & $0.99$ & $0.99$ \\
ViT         & $0.58$ & $0.56$ & $0.56$ & $0.52$ \\

\bottomrule
\end{tabular}
\caption{Average cosine similarities for abstract (A) and concrete (C) concepts for the Bing-25 and YFCC-25 datasets.}
\label{tab:cosine_values}
\end{table}

Table~\ref{tab:cosine_values} shows a comparison of cosine similarity scores for the top 25 nearest neighbors of an image, evaluated across different visual features. The similarity scores are generally consistent across feature type both for concrete and abstract targets, and across different datasets. Vision Transformer (ViT) stand out for having lower scores compared to the other features.

\subsection{Crowd-sourcing Collections}

\begin{figure*}[!thb]
	\center
    \includegraphics[width=0.97\textwidth]{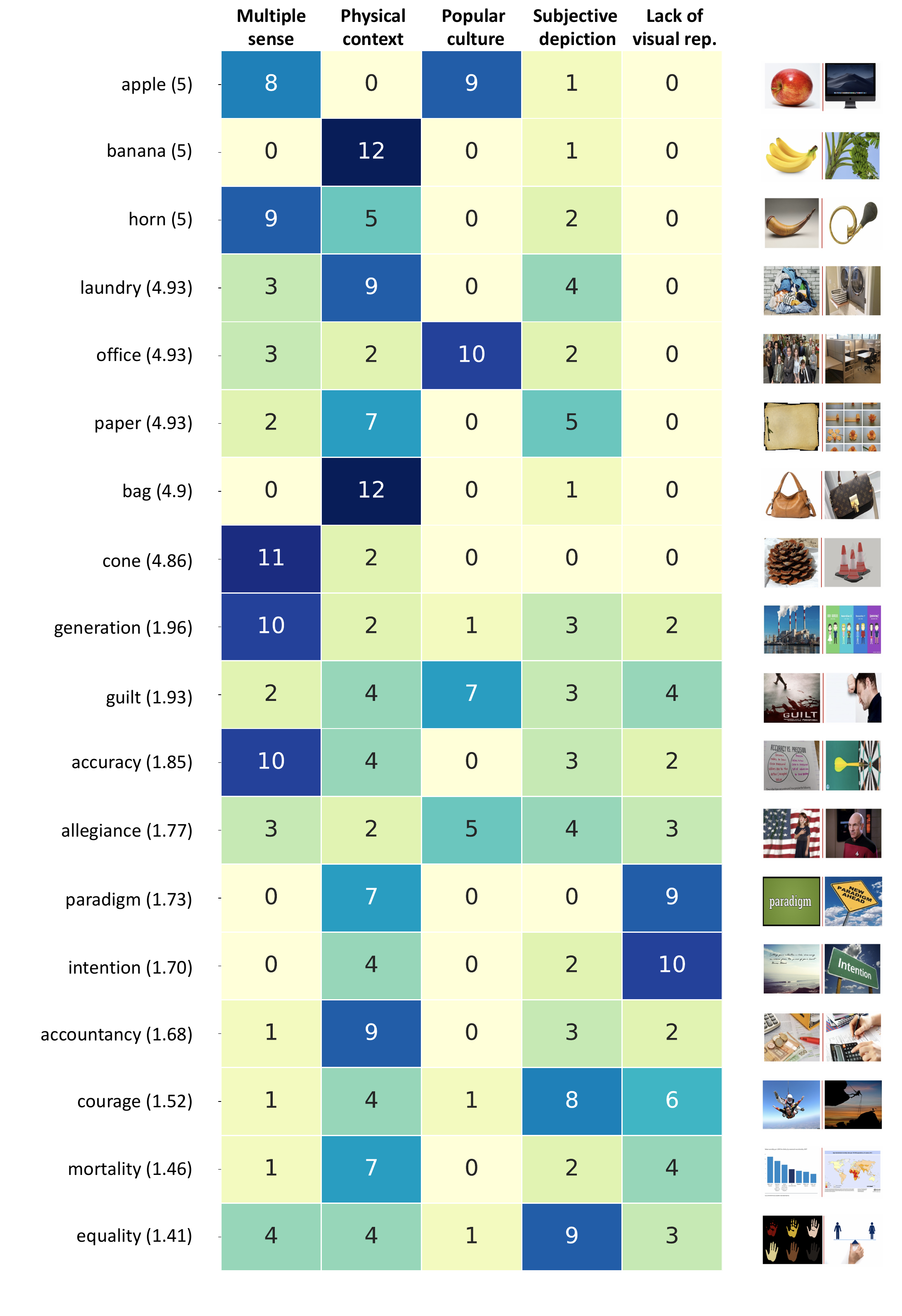}
    
	\caption{Main reasons of visual diversity between two images of $18$ concepts ($8$ concrete and $10$ abstract) according to $13$ participants. At least one reason had to be selected for each concept.}
	\label{imsstudy}
\end{figure*}

As discussed in Section~\ref{sec:study3}, we collected data using crowd-sourcing methods. The classification of $18$ concepts ($8$ concrete and $10$ abstract) in five ``reasons for visual diversity" is reported in Figure~\ref{imsstudy}. 

\noindent Tables~\ref{tab:amt_annotations_abs} and \ref{tab:amt_annotations_conc} provide examples of words describing the images of five concrete and five abstract concepts.

\subsection{Model Details}

In Study~1, we used three classifiers: Random forest (RF), SVM and Logistic Regression from the scikit-learn library \cite{scikit-learn}, and performed an extensive hyper-parameter search with $5$-fold cross-validation. For RF, the hyper-parameters included \textit{number of estimators} (trees), \textit{max\_depth} (maximum depth of the tree), \textit{min\_samples\_split} (minimum number of samples required to split an internal node), \textit{min\_samples\_leaf} (minimum number of samples required at a leaf node) and \textit{max\_features} (number of features to be considered for determining the best split). 

For feature extraction for the YOLO model, we used an NVIDIA RTX A6000 GPU. It takes around $8$ hours of GPU processing to extract YOLO features. The computation of nearest neighbors takes multiple weeks.

\begin{table*}[!tbh]
\centering
\setlength\tabcolsep{12pt} % Adjust the horizontal padding
\begin{tabular}{lrrrrrrrr}
\toprule
 & 
  \multicolumn{2}{c|}{ \textbf{YFCC-100}} &
  \multicolumn{2}{c|}{ \textbf{YFCC-200}} &
  \multicolumn{2}{c|}{ \textbf{YFCC-300}} &
  \multicolumn{2}{c}{  \textbf{YFCC-400}}  \\ \cmidrule(lr){2-9} 
\multirow{-2}{*}{\textbf{Attribute}} &
  \multicolumn{1}{c}{ \textbf{A}} &
   \multicolumn{1}{c}{\textbf{C}} &
   \multicolumn{1}{c}{ \textbf{A}} &
   \multicolumn{1}{c}{\textbf{C}} &
  \multicolumn{1}{c}{ \textbf{A}} &
   \multicolumn{1}{c}{\textbf{C}} &
  \multicolumn{1}{c}{ \textbf{A}} &
   \multicolumn{1}{c}{\textbf{C}} \\ \midrule

Color       & $1.28$ &	$0.79$ &	$0.99$ &	$0.72$ &	$0.88$ &	$0.68$ &	$0.86$ &	$0.66$ \\ 
HOG         & $0.47$ &	$0.48$ &	$0.34$ &	$0.46$ &	$0.32$ &	$0.45$ &	$0.34$ &	$0.44$ \\ 
Texture     & $0.30$ &	$0.24$ &	$0.26$ &	$0.25$ &	$0.26$ &	$0.26$ &	$0.27$ &	$0.26$ \\ 
GIST        & $0.69$ &	$0.61$ &	$0.53$ &	$0.58$ &	$0.50$ &	$0.57$ &	$0.51$ &	$0.56$ \\ 
SURF        & $0.65$ &	$0.55$ &	$0.44$ &	$0.53$ &	$0.40$ &	$0.53$ &	$0.52$ &	$0.52$ \\ 
YOLO        & $1.58$ &	$1.38$ &	$1.70$ &	$1.46$ &	$1.65$ &	$1.50$ &	$1.66$ &	$1.54$ \\ 
Object Loc. & $0.20$ &	$0.23$ &	$0.19$ &	$0.24$ &	$0.21$ &	$0.25$ &	$0.23$ &	$0.26$ \\ \midrule
Combined    & $1.03$ &	$0.85$ &	$0.78$ &	$0.80$ &	$0.71$ &	$0.76$ &	$0.70$ &	$0.76$ \\ \midrule
Simclr      & $0.80$ &	$0.65$ &	$1.67$ &	$1.67$ &	$1.40$ &	$1.45$ &	$0.53$ &	$0.56$ \\ 
ViT         & $2.79$ &	$6.71$ &	$2.30$ &	$6.67$ &	$4.55$ &	$11.99$ &	$2.26$ &	$6.55$ \\ 
\bottomrule
\end{tabular}
\caption{Average percentage of visual nearest neighbors
(out of 100, 200, 300 or 400, respectively) associated with the same
abstract (A) or concrete (C) concept.} 
\vspace{+5mm}
\label{tab:results_nn_variable_images}
\end{table*}

\begin{table*}[!tbh]
\centering
\footnotesize
\begin{tabular}{|p{2.2cm}|p{1.5cm}|p{1.5cm}|p{1.5cm}|p{1.5cm}|}

\hline
 &
  \multicolumn{2}{l|}{\hfil \textbf{Bing}} &
  \multicolumn{2}{l|}{\hfil \textbf{YFCC}} \\ \cline{2-5} 
\multirow{-2}{*}{\textbf{Visual Attribute}} & 
  \multicolumn{1}{l|}{\hfil \textbf{$\rho$}} &
  \hfil \textbf{RMSE} &
  \multicolumn{1}{l|}{\hfil \textbf{$\rho$}} &
  \hfil \textbf{RMSE} \\ \hline
  Color       & \hfil  $0.52$ & \hfil $1.34$ & \hfil $0.16$ & \hfil $1.58$ \\ \hline
  HOG           & \hfil $0.24$ & \hfil $1.53$ & \hfil $0.12$  & \hfil $1.60$  \\ \hline
  Texture     & \hfil $0.42$ & \hfil $1.41$ & \hfil $0.17$  & \hfil $1.57$ \\ \hline
  GIST         & \hfil $0.38$ & \hfil $1.43$ & \hfil $0.07$  & \hfil $1.61$ \\ \hline
  SURF         & \hfil $0.49$ & \hfil $1.34$ & \hfil $0.07$ & \hfil $1.61$ \\ \hline
  YOLO        & \hfil $0.26$ & \hfil $1.54$ & \hfil $0.07$ & \hfil $1.61$ \\ \hline
  Object Location      & \hfil $0.21$  & \hfil $1.67$ & \hfil $0.01$ & \hfil $1.62$ \\ \hline\hline
  Combined    & \hfil $\textbf{0.63}$ & \hfil $\textbf{1.12}$ & \hfil $\textbf{0.30}$ & \hfil $\textbf{1.51}$ \\ \hline\hline
  SimClr       & \hfil $0.28$ &   \hfil $1.87$ & \hfil $0.17$ & \hfil $1.90$ \\ \hline
  ViT          & \hfil $0.56$ & \hfil $1.27$ & \hfil $0.20$ & \hfil $1.85$  \\ \hline
\end{tabular}
\caption{Spearman correlation scores ($\rho$) and Root-mean-squared-error (RMSE) comparing the predicted concreteness scores using different visual attributes to the \textit{Brysbaert} norms. Results for the Bing and the YFCC datasets. In bold-font we highlight the highest scores for each dataset.}
\vspace{+2mm}
\label{tab:corrFeatures}
\end{table*}

\begin{table*}[!tbh]
\centering
\begin{tabular}{|m{1.8cm}|m{2.2cm}|m{2.2cm}|m{2.2cm}|m{2.2cm}|m{2.2cm}|}

\hline
& \multicolumn{1}{c|}{\textbf{equality}} & \multicolumn{1}{c|}{\textbf{mortality}} & \multicolumn{1}{c|}{\textbf{courage}} & \multicolumn{1}{c|}{\textbf{accountancy}} & \multicolumn{1}{c|}{\textbf{intention}} \\ 
& \multicolumn{1}{c|}{\textbf{($1.41$)}} & \multicolumn{1}{c|}{\textbf{($1.46$)}} & \multicolumn{1}{c|}{\textbf{($1.52$)}} & \multicolumn{1}{c|}{\textbf{($1.68$)}} & \multicolumn{1}{c|}{\textbf{($1.70$)}} \\ 
\hline &

\vspace{2pt}\raisebox{-0.45\height}{\centering \includegraphics[width=.136\textwidth, height=2cm]{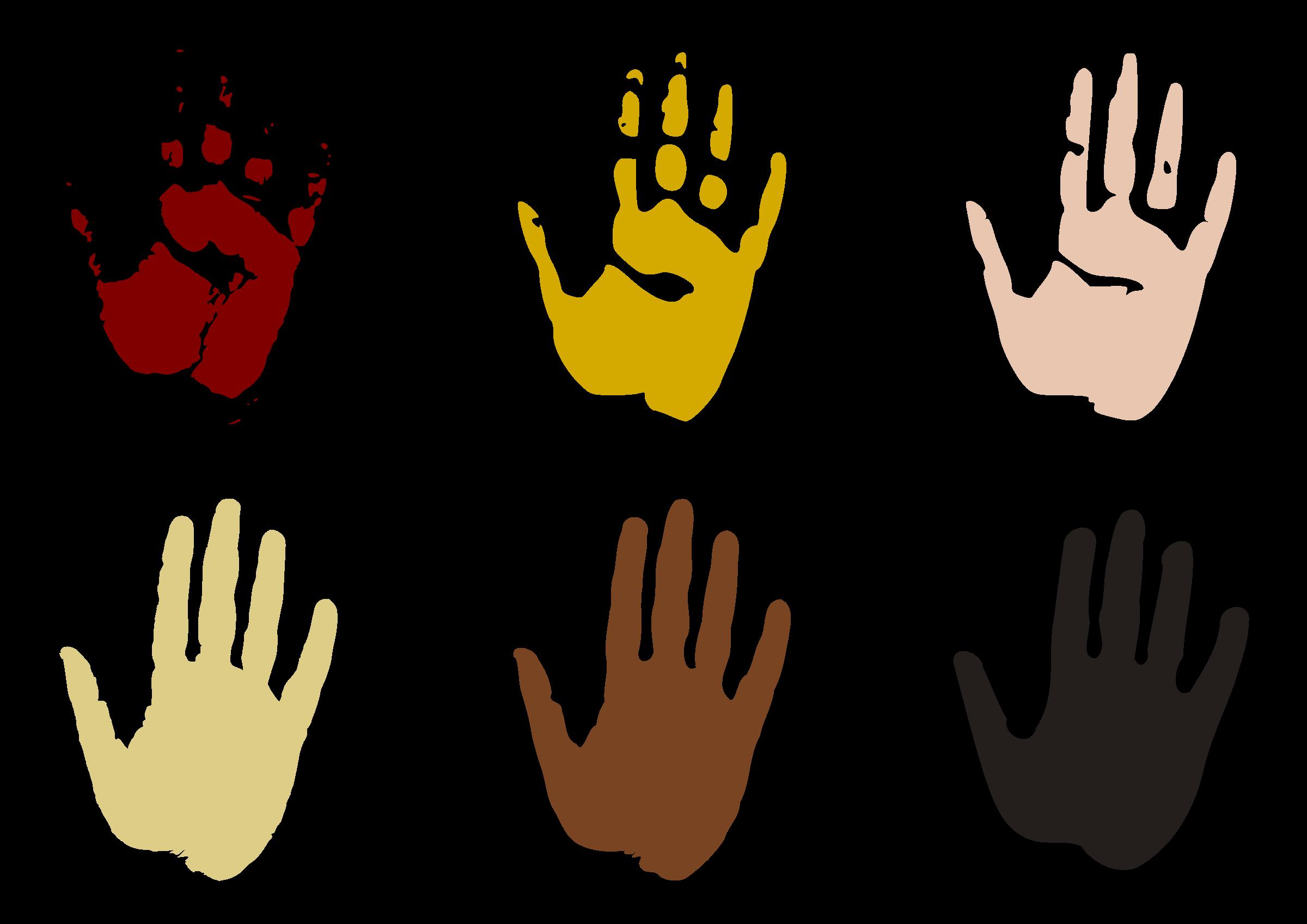}}\vspace{2pt}
 & \vspace{2pt}\raisebox{-0.45\height}{\centering \includegraphics[width=.136\textwidth, height=2cm]{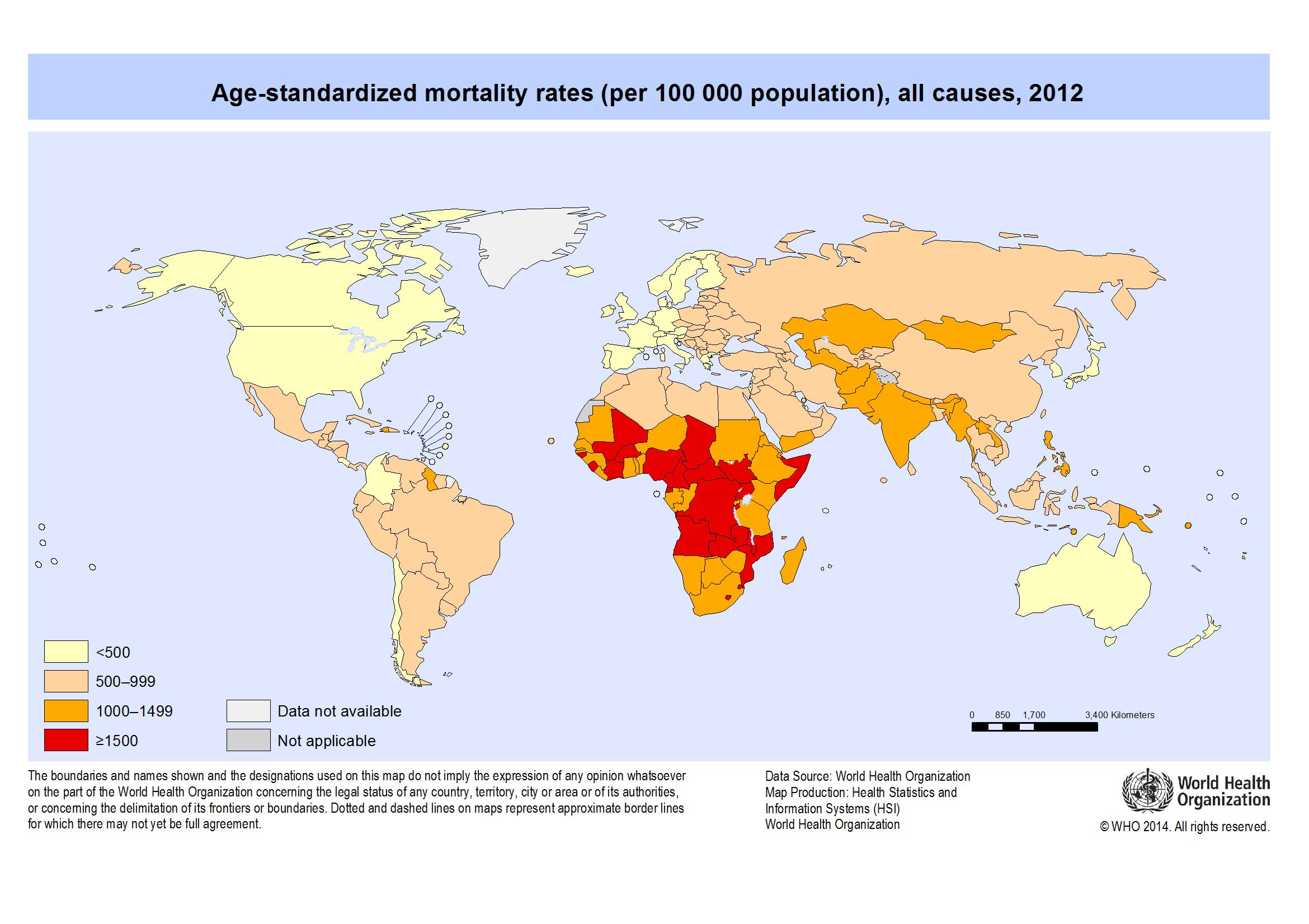}}\vspace{2pt}
 & \vspace{2pt}\raisebox{-0.45\height}{\centering \includegraphics[width=.136\textwidth, height=2cm]{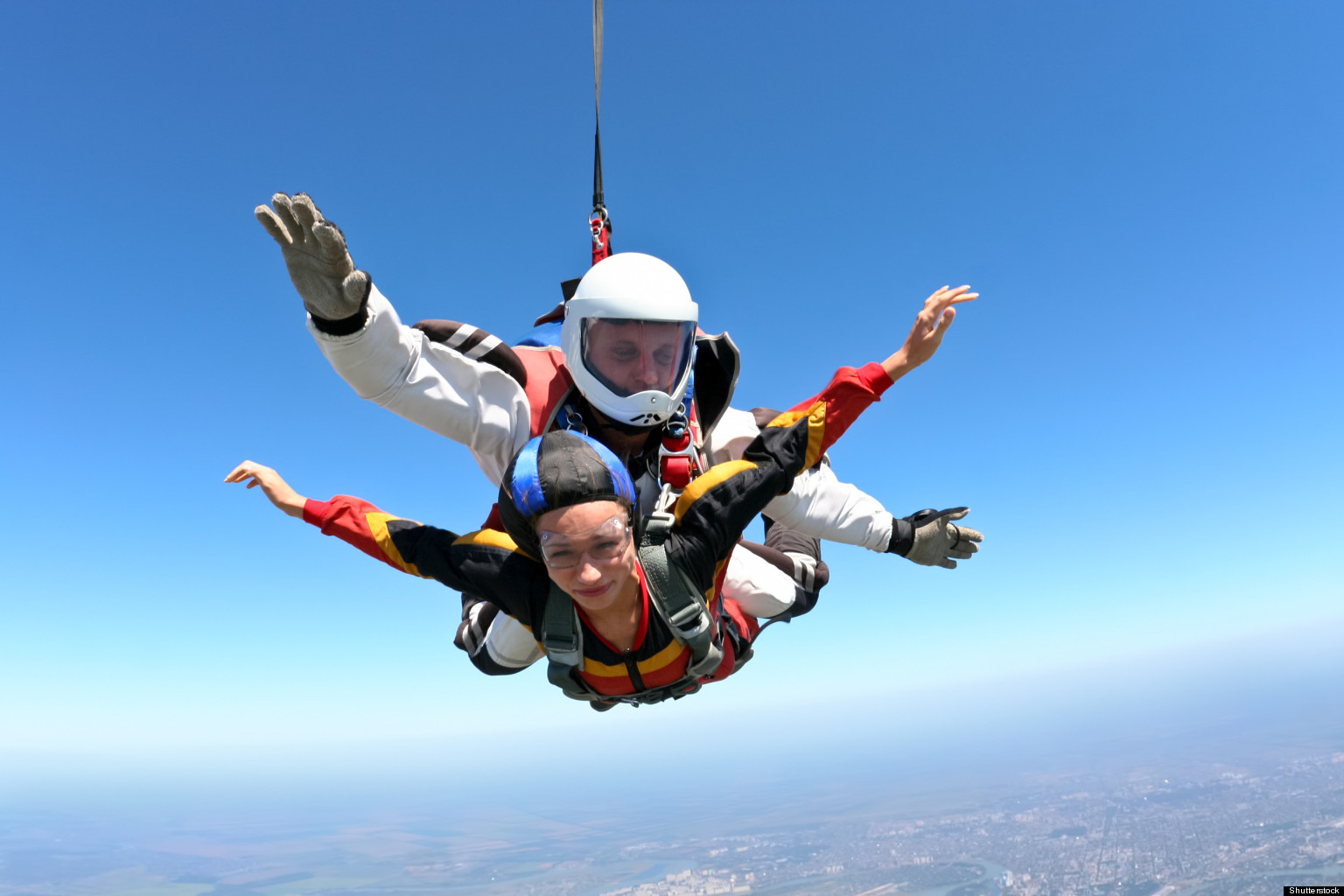}}\vspace{2pt}
 & \vspace{2pt}\raisebox{-0.45\height}{\centering \includegraphics[width=.136\textwidth, height=2cm]{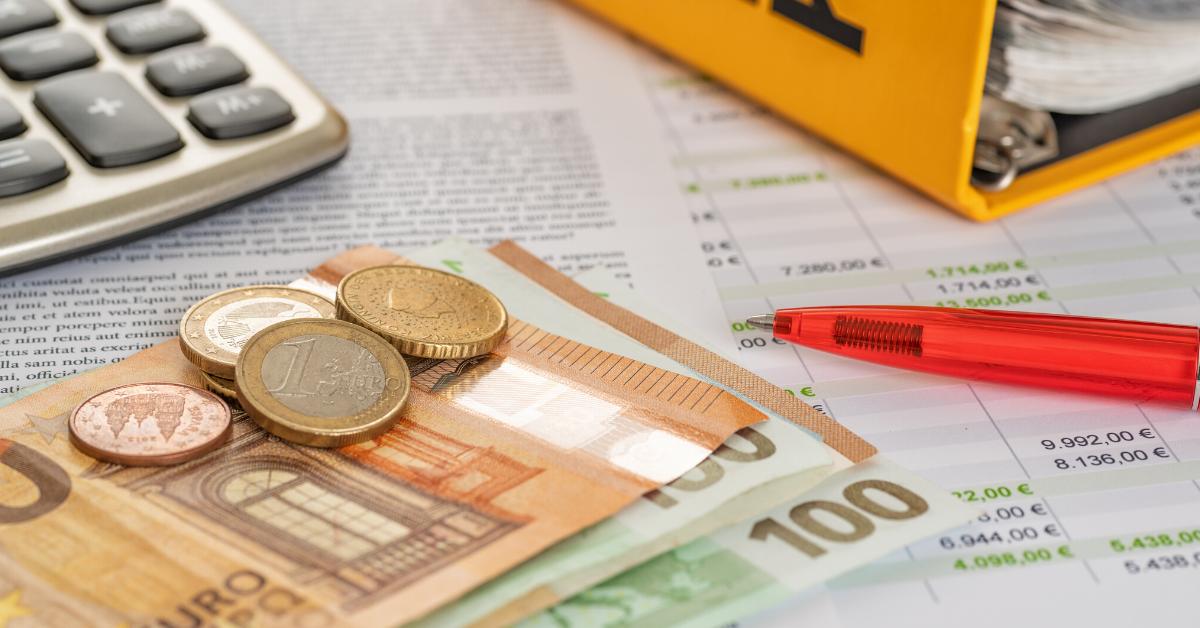}}\vspace{2pt}
 & \vspace{2pt}\raisebox{-0.45\height}{\centering \includegraphics[width=.136\textwidth, height=2cm]{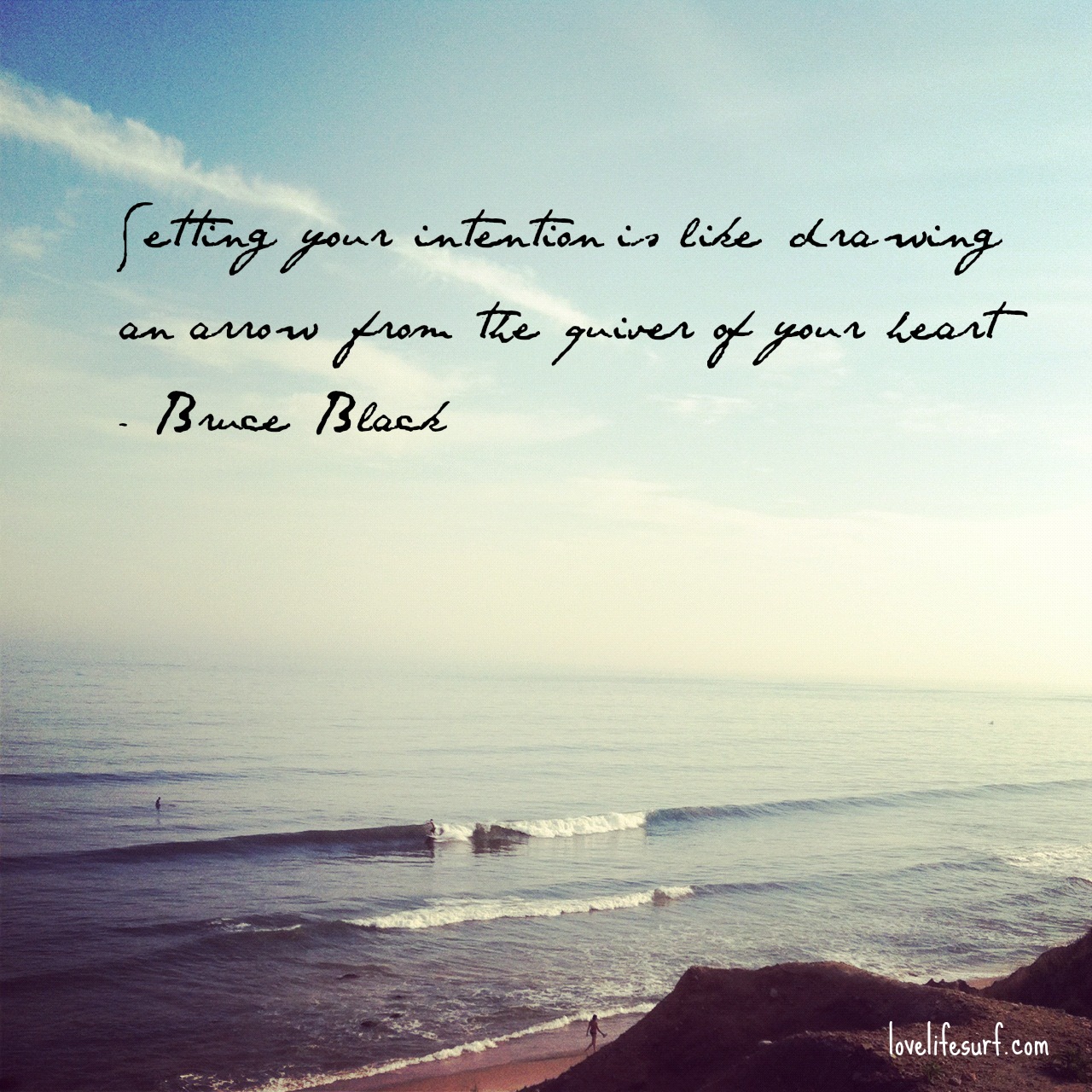}}\vspace{2pt}
 \\ \hline

\textbf{number of distinct annotations} &\multicolumn{1}{c|}{\multirow{1}{*}{$17$}} & \multicolumn{1}{c|}{\multirow{1}{*}{$18$}} & \multicolumn{1}{c|}{\multirow{1}{*}{$24$}} & \multicolumn{1}{c|}{\multirow{1}{*}{$10$}} &  \multicolumn{1}{c|}{\multirow{1}{*}{$18$}} \\ \hline

\textbf{Annotations} & \begin{tabular}[t]{@{}b{2cm}@{}}
red: $7$ \\
yellow: $7$ \\
brown: $5$ \\
hand: $3$ \\
grey: $2$ \\
pink: $2$ \\
black: $2$ \\
sandal: $2$ \\
hi: $1$ \\
ash: $1$ \\
orange: $1$ \\
white: $1$ \\
hand print: $1$ \\
color: $1$ \\
fingers: $1$ \\
six hand: $1$ \\
six colors: $1$ \\
\end{tabular}
& \begin{tabular}[t]{@{}l@{}} map: $6$ \\ world map: $4$ \\ sea: $4$ \\ country: $3$ \\ continent: $3$ \\ ocean: $2$ \\ yellow: $2$ \\ earth: $2$ \\ orange: $1$ \\ india: $1$ \\ world: $1$ \\ desert: $1$ \\ articles: $1$ \\ red: $1$ \\ letters: $1$ \\ lands: $1$ \\ mortality rate:$1$ \\ population: $1$ \end{tabular} & 
\begin{tabular}[t]{@{}l@{}} sky: $6$ \\ fly: $6$ \\ adventures: $3$ \\ diving: $2$ \\ exciting: $2$ \\ air: $2$ \\ helmet: $2$ \\ person: $2$ \\ man: $1$ \\ women: $1$ \\ skydive: $1$ \\ nature: $1$ \\ coat: $1$ \\ rope: $1$ \\ hand: $1$ \\ focus: $1$ \\ two: $1$ \\ male: $1$ \\ female: $1$ \\ advancer: $1$ \\ skydress: $1$ \\ flying: $1$ \\ hanging: $1$ \\ helpmate: $1$ \end{tabular}  &  
\begin{tabular}[t]{@{}l@{}} coin: $9$ \\ pen: $8$ \\ calculator: $8$ \\ money: $5$ \\ file: $5$ \\ calculate: $1$ \\ rupees: $1$ \\ pencil: $1$ \\ paper: $1$ \\ euro notes: $1$ \end{tabular} & 
\begin{tabular}[t]{@{}l@{}} sky: $6$ \\ sea: $5$ \\ beach: $3$ \\ waves: $3$ \\ sand: $3$ \\ quotes: $2$ \\ water: $2$ \\ stone: $1$ \\ white: $1$ \\ motivation: $1$ \\ happy life: $1$ \\ good \\ \hspace*{+2mm}intention:$1$ \\ peaceful: $1$ \\ set goal: $1$ \\ blue: $1$ \\ post card: $1$ \\ ocean: $1$ \\ words: $1$  \end{tabular} \\ \hline
\end{tabular}
\caption{Words generated by nine participants when answering to the question ``What is depicted in each image?". Examples for  five images of \textit{abstract concepts} (and their concreteness score).}
\label{tab:amt_annotations_abs}
\end{table*}

\begin{table*}[!tbh]
\centering
\begin{tabular}{|m{1.8cm}|m{2.2cm}|m{2.2cm}|m{2.2cm}|m{2.2cm}|m{2.2cm}|}

\hline
& \multicolumn{1}{c|}{\textbf{office}} & \multicolumn{1}{c|}{\textbf{laundry}} & \multicolumn{1}{c|}{\textbf{horn}} & \multicolumn{1}{c|}{\textbf{banana}} & \multicolumn{1}{c|}{\textbf{apple}} \\ 
& \multicolumn{1}{c|}{\textbf{($4.93$)}} & \multicolumn{1}{c|}{\textbf{($4.93$)}} & \multicolumn{1}{c|}{\textbf{($5.00$)}} & \multicolumn{1}{c|}{\textbf{($5.00$)}} & \multicolumn{1}{c|}{\textbf{($5.00$)}} \\ 
\hline &

  \vspace{2pt}\raisebox{-0.45\height}{\centering \includegraphics[width=.136\textwidth, height=2cm]{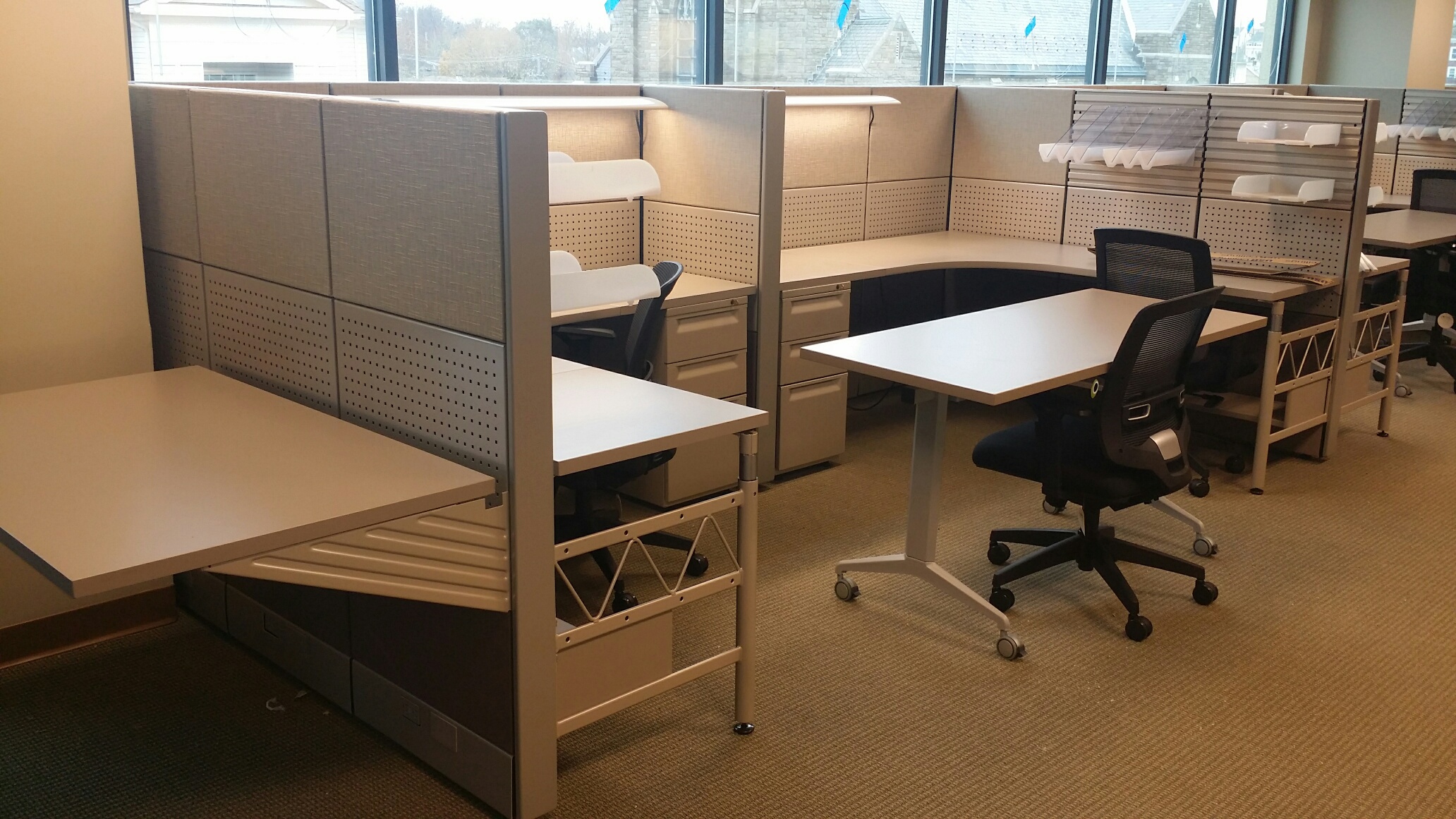}}\vspace{2pt}
 & \vspace{2pt}\raisebox{-0.45\height}{\centering \includegraphics[width=.136\textwidth, height=2cm]{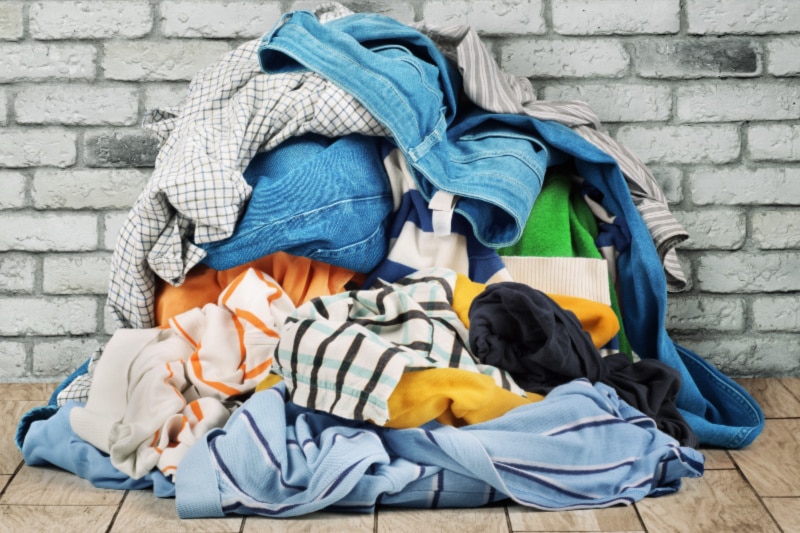}}\vspace{2pt}
 & \vspace{2pt}\raisebox{-0.45\height}{\centering \includegraphics[width=.136\textwidth, height=2cm]{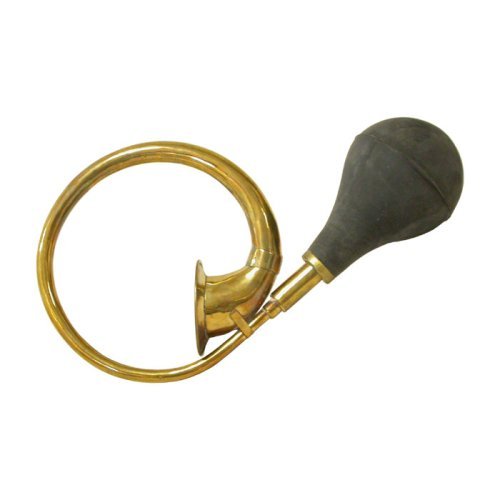}}\vspace{2pt}
 & \vspace{2pt}\raisebox{-0.45\height}{\centering \includegraphics[width=.136\textwidth, height=2cm]{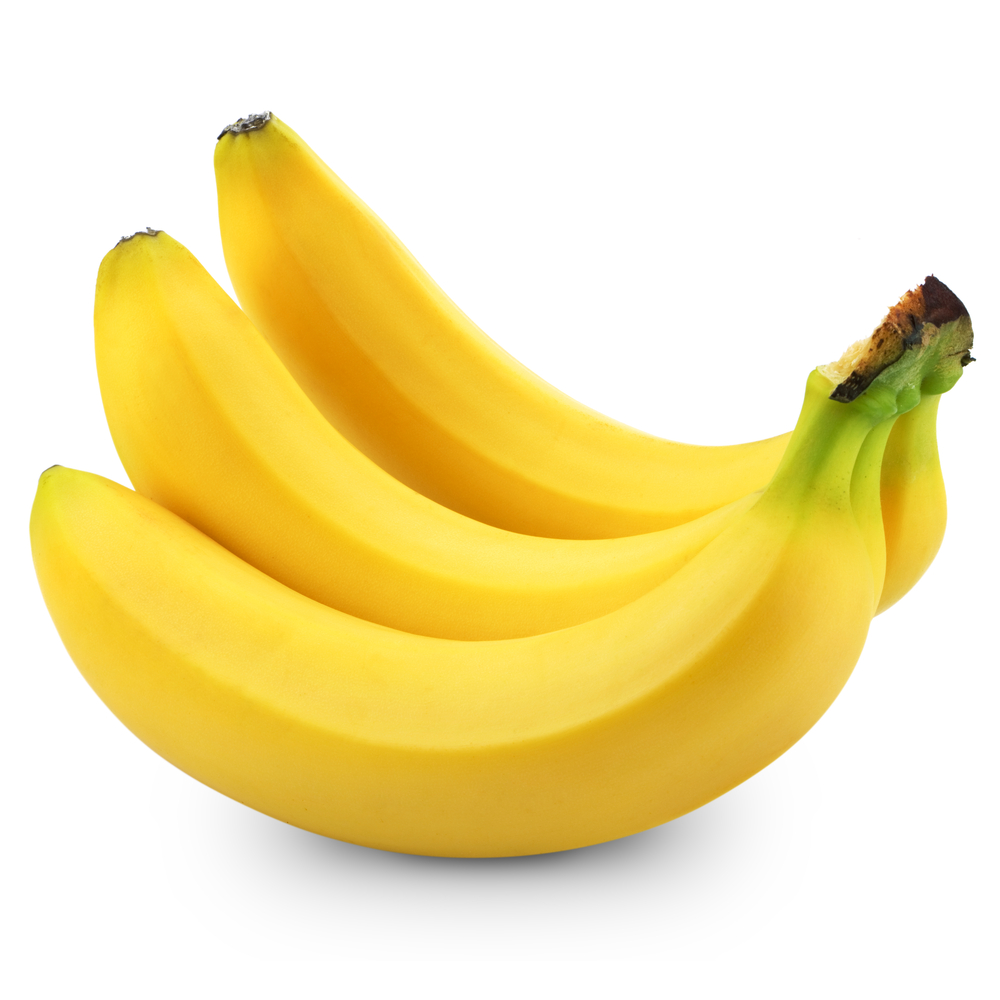}}\vspace{2pt}
 & \vspace{2pt}\raisebox{-0.45\height}{\centering \includegraphics[width=.136\textwidth, height=2cm]{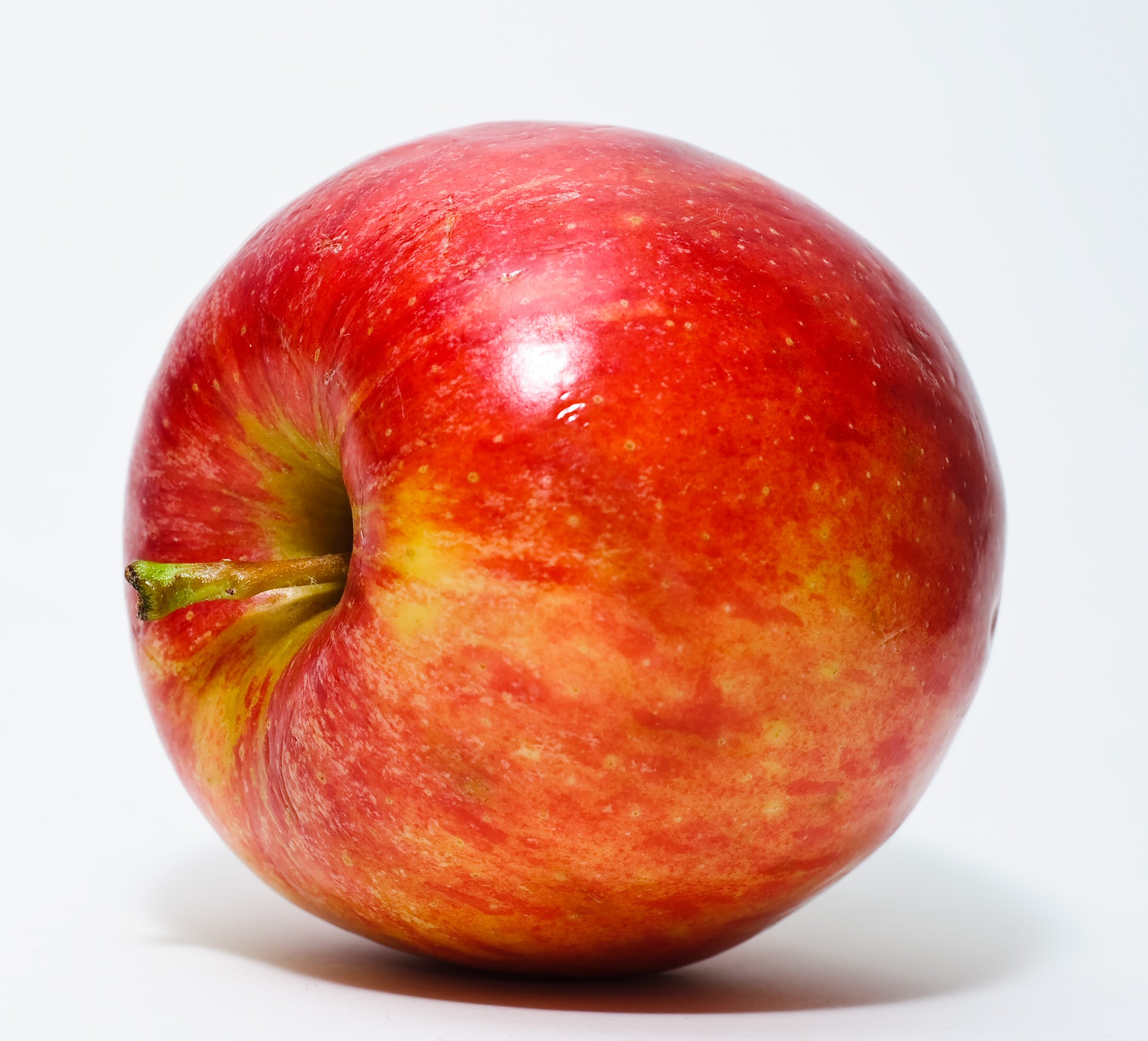}}\vspace{2pt}
 \\ \hline

\textbf{number of distinct annotations} &\multicolumn{1}{c|}{\multirow{1}{*}{$14$}} & \multicolumn{1}{c|}{\multirow{1}{*}{$16$}} & \multicolumn{1}{c|}{\multirow{1}{*}{$29$}} & \multicolumn{1}{c|}{\multirow{1}{*}{$14$}} &  \multicolumn{1}{c|}{\multirow{1}{*}{$16$}} \\ \hline
\textbf{Annotations} & \begin{tabular}[t]{@{}b{2cm}@{}}
chair: $9$\\ table: $8$\\ window: $7$\\ desk: $3$\\ room: $2$\\ glass: $2$\\ light: $1$\\ furniture: $1$\\ office \\ \hspace*{+2mm}furniture: $1$\\ building: $1$\\ cotton: $1$\\ floor: $1$\\ office: $1$\\ drawer: $1$ \\ 
\end{tabular}
& \begin{tabular}[t]{@{}l@{}} clothes: $8$\\ wall: $6$\\ pant: $4$\\ jeans: $3$\\ laundry: $2$\\ dress: $2$\\ shirt: $2$\\ floor: $2$\\ bricks: $2$ \\  color: $2$\\ 
 garments: $1$\\ trousers: $1$\\ stone: $1$\\ tiles: $1$\\ tshirt: $1$\\ blue: $1$  \end{tabular} & 
\begin{tabular}[t]{@{}l@{}} horn: $5$ \\ brass: $3$ \\ retro old- \\ \hspace*{+2mm}timer: $1$ \\ brass bulb: $1$ \\ motor horn: $1$ \\ rubber horn: $1$ \\ steel: $1$ \\ rubber: $1$ \\ circle: $1$ \\ oval: $1$ \\ honking \\ \hspace*{+2mm}sound : $1$ \\ brass honking \\ \hspace*{+2mm}instrument: $1$ \\ sound: $1$ \\ metal: $1$ \\ mike: $1$ \\ black: $1$ \\ rubber bulb: $1$ \\ musical \\ \hspace*{+2mm}instrument: $1$ \\ sound \\ \hspace*{+2mm}instrument: $1$ \\ signal horn: $1$ \\ military \\ \hspace*{+2mm}bugle: $1$ \\ brass \\ \hspace*{+2mm}instrument: $1$ \\ hunting horn:$1$ \\ conical horn: $1$ \\ honk: $1$ \\ rubber top: $1$ \\ metal \\ \hspace*{+2mm}instrument: $1$ \\ bulb horn: $1$ \\large circular: $1$ \end{tabular}  &  
\begin{tabular}[t]{@{}l@{}} banana: $13$\\ yellow: $8$\\ three: $3$\\ fruit: $2$\\ three \\ \hspace*{+2mm}banana: $1$\\ fresh fruit: $1$\\ very sweet \\\hspace*{+2mm}fruit: $1$\\ white: $1$\\ green: $1$\\ fresh: $1$\\ sweet: $1$\\ healthy: $1$\\ curved: $1$\\ ripened: $1$ \end{tabular} & 
\begin{tabular}[t]{@{}l@{}} fruit: $7$\\ apple: $6$\\ fresh: $4$\\ red apple: $3$\\ red: $3$\\ stem: $2$\\ health: $2$\\ eating: $1$\\ good for \\health: $1$\\ organic: $1$\\ one: $1$\\ fruits: $1$\\ fresh fruits: $1$\\ paradise \\ \hspace*{+2mm}apple: $1$\\ shadow: $1$\\ healthy: $1$  \end{tabular} \\ \hline
\end{tabular}
\caption{Words generated by nine participants when answering to the question ``What is depicted in each image?". Examples for  five images of \textit{concrete concepts} (and their concreteness score).}
\label{tab:amt_annotations_conc}
\end{table*}

\subsection{Regression Analysis}

We use a \textit{Gradient Boosted trees} model 
to predict the concreteness of each target concept using the eigenvalues of the combined visual features described in Section~\ref{sec:features} as predictors. The predicted concreteness scores are compared against the \textit{Brysbaert} norms using Spearman’s rank-order correlation coefficient $\rho$. We use an $80$:$20$ data split between  train and test sets with Monte Carlo cross-validation. 

As shown in Table~\ref{tab:corrFeatures}, the combination of all the low-level features (Combined) achieves the highest results for both datasets and outperforms both ViT and SimClr more complex representations. This is in line with the classification results. Similar to classification, we also further investigate the sampling bias of images, we conduct similar analysis for concepts with $100,200,300,400$ and $500$ images. We see similar results as depicted in Figure~\ref{classification_classwise} in the main text. As expected, Spearman correlations generally improve with the inclusion of more images, as increased data helps to average out noise.

\clearpage
\includepdf[pages=-]{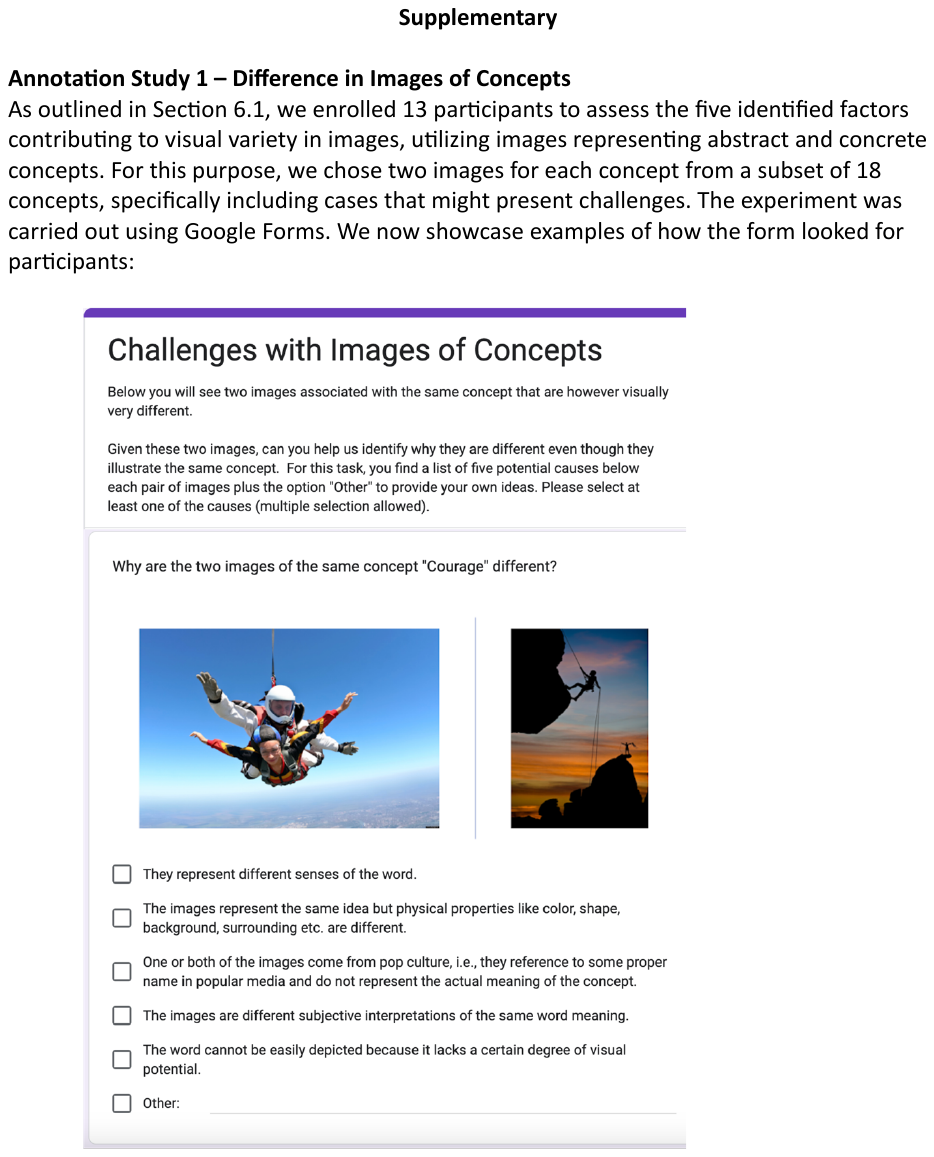}

\end{document}